\documentclass{article}

\usepackage{hyperref}
\usepackage{url}
\hypersetup{
    citecolor=CadetBlue,
    colorlinks=true,
    linkcolor=CadetBlue,
    filecolor=magenta,      
    urlcolor=cyan}

\usepackage{graphicx}
\usepackage{booktabs} 
\usepackage{multirow}
\usepackage{xcolor}
\definecolor{CadetBlue}{RGB}{95,158,160}
\usepackage[frozencache,cachedir=_minted]{minted}
\usepackage{placeins}

\usepackage{mathtools}
\usepackage{subcaption}

\usepackage[most]{tcolorbox}
\tcbuselibrary{minted, breakable}

\newtcblisting[auto counter, list inside=mypython]{pythoncode}[2][]{
    listing engine=minted,
    minted language=python,
    minted options={
        linenos,
        breaklines,
        tabsize=4
    },
    colback=lightgray!20,  
    colframe=black,        
    toprule=1pt,           
    bottomrule=1pt,        
    leftrule=0pt,          
    rightrule=0pt,         
    sharp corners,         
    breakable,             
    listing only,          
    title={Algorithm \thetcbcounter: #2}, 
    #1                     
}

\newcounter{pseudocode}
\newenvironment{pseudocodebox}[2]{
    \refstepcounter{pseudocode}
    \begin{tcolorbox}[
        breakable,
        colback=lightgray!10,
        colframe=black,
        toprule=1pt,
        bottomrule=1pt,
        leftrule=0pt,
        rightrule=0pt,
        sharp corners,
        title={Pseudocode \thepseudocode: #2}
    ]
    \label{#1}
}{
    \end{tcolorbox}
}

\usepackage{xspace}
\usepackage{makecell}
\newcommand{\method}[1]{\texttt{#1}\xspace}

\newcommand{\FIMzeroshot}{\method{FIM-PP}(zs)\xspace}
\newcommand{\FIMfine}{\method{FIM-PP}(f)\xspace}

\newcommand{\dataset}[1]{\textsc{#1}}
\newcommand{\datasetbf}[1]{\textbf{\dataset{#1}}}

\usepackage{amssymb}

\newcommand{\tinymath}[1]{\text{\tiny{\(#1\)}}}
\newcommand{\nomath}[1]{}

\newcommand{\OTD}{OTD\xspace}
\newcommand{\RMSEe}{\text{RMSE\textsubscript{$e$}}\xspace}
\newcommand{\RMSEdt}{\text{RMSE\textsubscript{$\Delta t$}}\xspace}
\newcommand{\SMAPEdt}{\text{sMAPE\textsubscript{$\Delta t$}}\xspace}
\newcommand{\accuracy}{Acc\xspace}
\newcommand{\ours}{\texttt{EVIL}\xspace}
\newcommand{\ourssynth}{\texttt{EVIL (synthetic prior)}\xspace}

 \usepackage[preprint]{neurips_2026}

\usepackage[utf8]{inputenc} 
\usepackage[T1]{fontenc}    
\usepackage{hyperref}       
\usepackage{url}            
\usepackage{booktabs}       
\usepackage{amsfonts}       
\usepackage{nicefrac}       
\usepackage{microtype}      
\usepackage{xcolor}         

\title{EVIL: Evolving Interpretable Algorithms for Zero-Shot Inference on Event Sequences and Time Series with LLMs}

\author{%
  David Berghaus\textsuperscript{1, 2}\\
Lamarr Institute\textsuperscript{1}, Fraunhofer IAIS\textsuperscript{2}\\
\small \texttt{david.berghaus@iais.fraunhofer.de} 
}

\begin{document}

\maketitle

\begin{abstract}
We introduce \ours (\textbf{EV}olving \textbf{I}nterpretable algorithms with \textbf{L}LMs), an approach that uses LLM-guided evolutionary search to discover simple, interpretable algorithms for dynamical systems inference.
Rather than training neural networks on large datasets, \ours evolves pure Python/NumPy programs that perform zero-shot, in-context inference across datasets.
We apply \ours to three distinct tasks: next-event prediction in temporal point processes, rate matrix estimation for Markov jump processes, and time series imputation.
In each case, a single evolved algorithm generalizes across all evaluation datasets without per-dataset training (analogous to an amortized inference model).
To the best of our knowledge, this is the first work to show that LLM-guided program evolution can discover a single compact inference function for these dynamical-systems problems.
Across the three domains, the discovered algorithms are often competitive with, and even outperform, state-of-the-art deep learning models while being orders of magnitudes faster, and remaining fully interpretable.
\end{abstract}

\section{Introduction}

Recent work on dynamical systems has been dominated by deep learning methods, including temporal point processes \cite{mei2017neural,mei2022transformer,cdiff}, Markov jump processes \cite{neural_mjp,fim_mjp}, and time series imputation \cite{tashiro2021csdi,seifner2025zeroshot}.
However, these approaches share several limitations:
(i) They require large curated training sets and are often expensive to train and run.
(ii) They produce opaque predictions, making it difficult to understand \emph{why} a particular prediction was made.
(iii) They often generalize poorly to distribution shifts between training and test data, or to output dimensions not seen during training.

We propose a fundamentally different strategy:
Instead of learning parameters of a neural network from data, we use LLM-guided evolutionary search to \emph{discover} simple, interpretable algorithms that solve these tasks directly.
Our approach, \ours (\textbf{EV}olving \textbf{I}nterpretable algorithms with \textbf{L}LMs), leverages AlphaEvolve-style program evolution \cite{novikov2025alphaevolve} to iteratively refine Python functions that take in context data and produce predictions using only NumPy operations.

Our main contributions are threefold.
First, we introduce a new use of LLM-guided program evolution: not to solve a single fixed instance or dataset, but to discover an amortized inference function that can be reused across datasets within a problem class and generalizes to arbitrary numbers of dimensions.
In our setting, the goal is to evolve one simple Python/NumPy procedure per task that takes context observations and directly produces predictions at test time, without per-dataset retraining or architectural changes.
To the best of our knowledge, this is the first use of AlphaEvolve-style \cite{novikov2025alphaevolve} program evolution in this way.
Second, we provide empirical evidence that such amortized functions actually exist for several nontrivial problem classes.
Across temporal point-process prediction, Markov jump process inference, and time-series imputation, the evolved programs generalize across datasets in a zero-shot, in-context manner and are often already competitive with substantially more complex learned baselines, while being faster, cheaper, and fully interpretable.
This is notable because it was not previously clear that such compact, reusable inference functions could be found at all across diverse datasets, nor that an LLM-guided evolutionary process could discover them automatically.
Third, we use these three applications to illustrate a broader picture: \ours is not a collection of task-specific tricks, but a general recipe for searching over simple inference procedures that exploit structure in contextual data.
Taken together, our results suggest that for at least some dynamical-systems inference problems, such discovered heuristics can offer a practical alternative to increasingly complex task-specific learned models.

\begin{figure}[t]
    \centering
    \includegraphics[width=\linewidth]{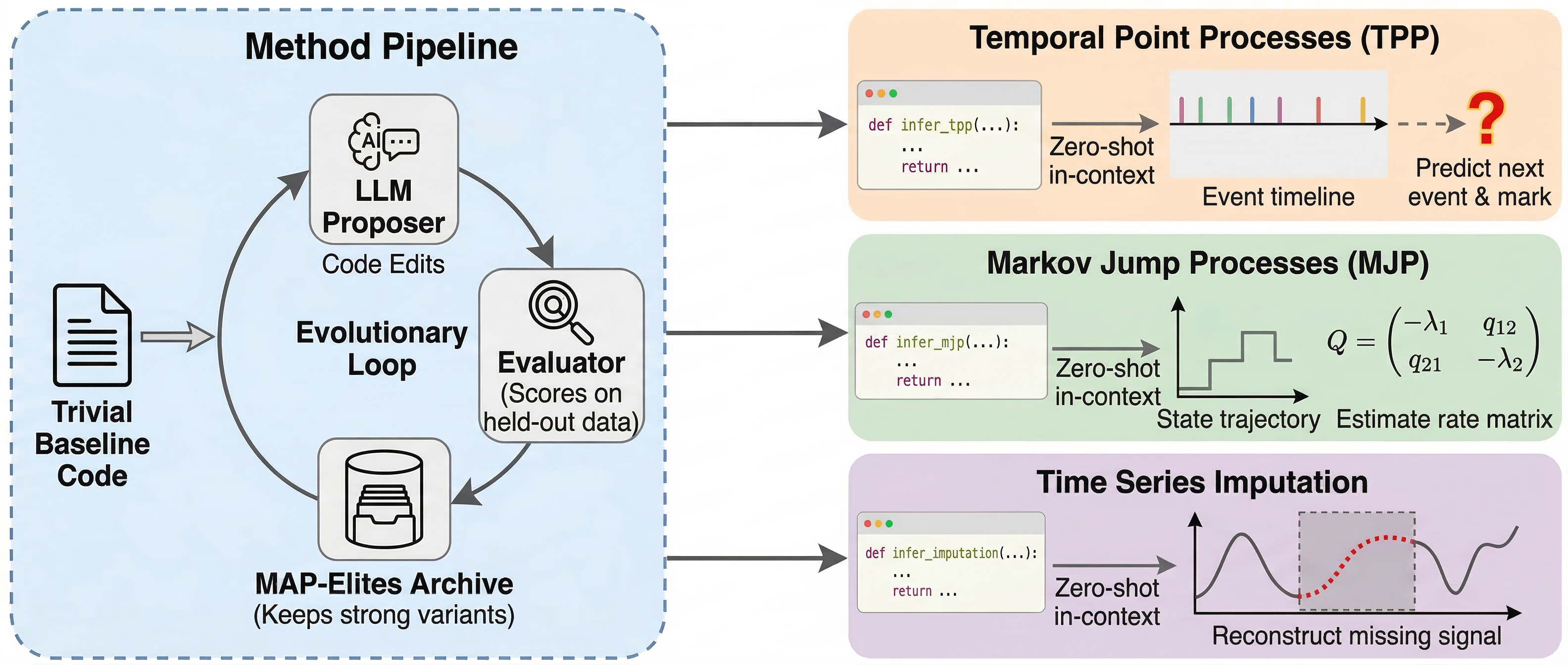}
    \caption{Overview of the \ours approach. The same EVIL evolutionary procedure is applied separately to temporal point processes, Markov jump processes, and time-series imputation, yielding one interpretable Python/NumPy inference function per task that generalizes across datasets in a zero-shot, in-context manner.}
    \label{fig:overview}
    \vspace{-8pt}
\end{figure}

The main practical advantages of \ours are:
\begin{itemize}
    \setlength\itemsep{0pt}
    \item \textbf{Competitive performance}: \ours is competitive with state-of-the-art deep learning methods across three distinct dynamical systems tasks, and outperforms them on several benchmarks (Sections~\ref{sec:tpp}--\ref{sec:imputation}).
    \item \textbf{Foundational generalization}: A single algorithm per task works across all datasets without per-dataset fine-tuning. Notably, and contrary to most deep learning methods, we found that \ours could even potentially generalize to unseen datasets with far more categories than encountered during evolution (Table~\ref{tab:mimic2}), and also behave more robust against distribution shifts (Figure~\ref{fig:dfr_entropy}).
    \item \textbf{Full interpretability}: Every evolved algorithm is a readable Python function whose logic can be inspected and understood, potentially providing more insight into the problem and confidence into the safety of the outputs (Appendix~\ref{app:algorithms}).
    \item \textbf{Simplicity}: The evolutions are straightforward to implement, enabling domain practitioners to implement and use the algorithms without having to become experts in deep learning. Moreover, contrary, to deep learning models, domain expert knowledge can easily be incorporated, either by direct modifications of the program, or by running additional evolutions with a more informed system prompt.
    \item \textbf{Rapid discovery}: Evolution takes a few minutes of wall-clock time and results in less than \$1 of API costs, compared to hours or days of neural network training.
    \item \textbf{Fast inference}: Predictions require only basic array operations, enabling real-time use on commodity hardware (Table~\ref{tab:tpp-runtime}).
    \item \textbf{Minimal data}: \ours can work with only small subsamples from existing training splits and requires significantly less data than neural baselines.
\end{itemize}

We hypothesize that \ours performs well for three reasons.
First, the LLM brings extensive prior knowledge about statistical methods and algorithm design, providing an informed starting point that neural networks must learn from scratch.
Second, evolved programs can perform direct and exact statistical computations on the context data rather than having to approximate these operations through learned parameters.
Third, the requirement that solutions be expressible as short programs imposes an implicit bias toward simpler approaches that generalize better, reminiscent of Occam's razor, similar to observations made in program synthesis for mathematical discovery \cite{georgiev2025mathematical}.
\footnote{Our source code is available at \url{https://anonymous.4open.science/r/evil_algorithms-374D/}.}

\section{Related Work}
\label{sec:related}

\subsection{LLM-Guided Program Evolution}
AlphaEvolve \cite{novikov2025alphaevolve} introduced an evolutionary framework where LLMs propose code modifications that are scored by automated evaluators, achieving advances in algorithm design and mathematical discovery \cite{georgiev2025mathematical}.
FunSearch \cite{romeraparedes2023mathematical} pioneered a similar approach for discovering novel solutions to combinatorial problems.
LLM-SR \cite{shojaee2024llmsr} applies LLM-based search to symbolic regression.

\subsection{Foundation Models and In-Context Inference}
Recent work on dynamical systems has explored synthetic pretraining and amortized in-context inference as a general alternative to fitting a new neural network for each dataset.
In the broader in-context-learning literature, prior-fitted neural networks learn to approximate predictive posteriors directly from context-target examples \cite{muller2022transformers}.
Within dynamical systems, the Foundation Inference Model (FIM) line instead emphasizes zero-shot estimation of system-defining operators or trajectories from contextual observations, including FIM-MJP for rate-matrix inference in Markov jump processes \cite{fim_mjp}, FIMs for inference of ordinary \cite{fim_ode} and stochastic \cite{fim_sde} differential equations, FIM-$\ell$ for zero-shot imputation in dynamical systems \cite{seifner2025zeroshot}, and FIM-PP for conditional-intensity inference in temporal point processes \cite{fim_pp}.
These works form the main foundation-model context for our setting; the domain-specific methods that require training on each individual dataset are introduced within the corresponding subsections in Sections~\ref{sec:tpp}--\ref{sec:imputation}.
Very recently, \citet{zhang2026context} proposed \emph{context parroting} (which is a simple baseline that copies relevant patterns from the input context) and show that it is surprisingly competitive with foundation models in chaotic time series forecasting, reinforcing the message that simple, interpretable baselines can challenge trained models.

\section{Methodology}
\label{sec:method}

\subsection{Problem Setup}
In each of our three applications, the task can be framed as: given a set of context observations and a target query, produce a prediction using a pure Python/NumPy function.
For point processes, this means predicting the next event given a history and context sequences.
For MJPs, it means estimating rate matrices from discrete observations.
For imputation, it means filling in missing values given partially observed time series.
In all cases, the function must work across datasets without modification---it operates in a zero-shot, in-context manner at test time.

\subsection{Evolutionary Search}
We use OpenEvolve \cite{sharma2025openevolve}, an open-source implementation of AlphaEvolve \cite{novikov2025alphaevolve}, to evolve our inference functions.
The evolution starts from a trivial initial program and iteratively improves it over the iterations.
At each iteration, an LLM proposes code modifications in diff format, the modified program is evaluated against a scoring function, and promising solutions are stored in an evolutionary database.
The database implements MAP-Elites \cite{mouret2015illuminating} with island-based populations to balance exploration and exploitation.
We use an ensemble of LLMs for code generation, with the system prompt constraining solutions to use only NumPy (no deep learning frameworks).
Full hyperparameters and system prompts are provided in Appendix~\ref{app:hyperparams}.
We kept the initial programs, system prompts, and scoring functions fixed throughout the study and did not optimize them based on benchmark results to avoid implicit overfitting.

\subsection{Evaluation Protocol}
To prevent overfitting to test data, we create our own evaluation subsplit from the existing training split of each dataset.
This means \ours has access to strictly less data than baseline methods that train on the full training set.
The scoring function for evolution measures prediction quality on this held-out subsplit (e.g., accuracy and RMSE for point processes, cross-entropy and RMSE for MJPs, MAE for imputation).
We report two variants: \ours, evolved on subsplits of the target datasets, and \ourssynth, evolved on synthetic data only.

To clarify what we mean by ``zero-shot,'' the final evolved program is applied at test time without any per-dataset training or parameter fitting. For \ours, the target-dataset subsplits are used only during evolution to guide the search toward a good general-purpose algorithm; the resulting program itself contains no dataset-specific parameters and generalizes to entirely unseen datasets (e.g., \dataset{MIMIC-II}, Table~\ref{tab:mimic2}).
\ourssynth provides the strictest zero-shot baseline, evolved entirely on synthetic data, to isolate whether an LLM-evolved heuristic can generalize from a broad synthetic prior to real-world datasets.

We keep the main experimental setup fixed to 100 LLM iterations, with 20\% of the suggestions coming from GPT-5 and 80\% from GPT-5-mini.
Additional ablations on dataset-specific evolution, variance across independent runs, and a substantially longer run are reported in Appendix~\ref{app:ablations}.

\section{Temporal Point Processes}
\label{sec:tpp}

\subsection{Background}
A marked temporal point process (MTPP) \cite{daley2003introduction} is a stochastic process generating sequences of events $(t_i, k_i)_{i=1}^{N}$, where each event has a timestamp $t_i \in \mathbb{R}^+$ and a categorical mark $k_i \in \{1, \dots, K\}$.
The process is characterized by its conditional intensity function $\lambda^*(t, k)$, governing the instantaneous rate of events given the history.
The central prediction task is forecasting the next $N$ events given a history of past events. Appendix~\ref{app:tpp_details} gives a slightly more detailed formal description, summarizes the datasets, and briefly reviews the point-process baselines.

\subsection{Experimental Setup}
We evolve a function that takes target histories and a pool of context sequences, and returns predicted next-event times and marks.
The initial program simply predicts the global median inter-event gap and the majority mark type.
The evolutions (100 iterations) took about 15 minutes wall-clock time.
We evolve two variants: \ours, which was evolved on small amounts of subdata from the target datasets in Table~\ref{tab:long-horizon-5-compact}, and \ourssynth, which was evolved only on subsplits of the synthetic data from \cite{fim_pp}.
This setup makes the comparison to FIM-PP \cite{fim_pp} particularly natural: like \ourssynth, FIM-PP is built around synthetic pretraining and is then evaluated either zero-shot or after finetuning on real data. At the same time, FIM-PP was pretrained on substantially more data overall (14.4 million events, whereas \ourssynth was evolved on only 72k).

\paragraph{Baselines.}
We compare against strong point-process baselines including A-NHP, NHP, IFTPP, Dual-TPP, CDiff, HawkesEM, and FIM-PP \cite{mei2017neural,mei2022transformer,shchur2020intensity,deshpande2021long,cdiff,hawkesem,fim_pp}.
These include neural architectures trained separately on each dataset, a classical Hawkes-process baseline, and a pretrained foundation model; all are substantially more complex than our final heuristic, and all except the pretrained FIM-PP variants require per-dataset training.

\subsection{Discovered Algorithm}
The evolved point-process heuristic (see Pseudocode~\ref{algpseud:tpp} and Python Algorithm~\ref{alg:tpp} in Appendix~\ref{app:algorithms}) can be summarized as:
\emph{(1) from the context sequences, estimate a smoothed mark-transition table and typical inter-event gaps, including mark-specific average gaps; (2) predict the next time by mixing recent gaps with the average gap associated with the last mark; (3) predict the next mark by combining local transition counts from the prefix with the global transition table.}

\subsection{Results and Analysis}
Table~\ref{tab:long-horizon-5-compact} summarizes long-horizon ($N{=}5$) prediction across five datasets.
While \ours achieves highly competitive predictive performance, the practical speedup is especially striking: for this $N{=}5$ benchmark, FIM-PP evaluation took 964s (about 16 minutes) on an A100 40GB GPU, whereas EVIL took 5s on a single Apple M3 CPU core, a nearly $200\times$ faster evaluation despite using much simpler hardware. The other neural baselines have even higher end-to-end cost because they must first be trained for each dataset.

\begin{table}[t]
\caption{Long-horizon prediction ($N{=}5$) on five real-world datasets. We show OTD and $\text{sMAPE}_{\Delta t}$ (both lower is better) for the best neural baseline (CDiff), FIM-PP variants, and \ours variants. Full results including all baselines are in Appendix~\ref{app:tpp_tables}. Baseline results from \cite{cdiff,fim_pp}. FIM(zs) is FIM-PP \cite{fim_pp} in zero-shot mode, FIM(f) is FIM-PP after finetuning on the target dataset. EVIL(s) is \ourssynth.}
\label{tab:long-horizon-5-compact}
\scriptsize
\centering
\setlength{\tabcolsep}{3pt}
\begin{tabular}{l cc cc cc cc cc}
\toprule
 & \multicolumn{2}{c}{\datasetbf{Taxi}} & \multicolumn{2}{c}{\datasetbf{Taobao}} & \multicolumn{2}{c}{\datasetbf{StackOv.}} & \multicolumn{2}{c}{\datasetbf{Amazon}} & \multicolumn{2}{c}{\datasetbf{Retweet}} \\
\cmidrule(lr){2-3}\cmidrule(lr){4-5}\cmidrule(lr){6-7}\cmidrule(lr){8-9}\cmidrule(lr){10-11}
Method & OTD & sM & OTD & sM & OTD & sM & OTD & sM & OTD & sM \\
\midrule
CDiff & $5.97$ & $89.5$ & $10.15$ & $124.3$ & $10.74$ & $100.6$ & $\mathbf{9.48}$ & $81.3$ & $15.86$ & $106.6$ \\
HawkesEM & $7.15$ & $86.4$ & $11.27$ & $164.1$ & $11.70$ & $\mathbf{82.8}$ & $14.64$ & $173.7$ & $15.69$ & $80.5$ \\
FIM(zs) & $6.77$ & $74.9$ & $15.95$ & $168.3$ & $11.52$ & $93.3$ & $11.12$ & $119.1$ & $15.75$ & $98.7$ \\
FIM(f) & $4.08$ & $71.1$ & $13.17$ & $146.9$ & $\mathbf{10.35}$ & $86.4$ & $10.03$ & $78.7$ & $15.65$ & $83.0$ \\
\midrule
\textbf{EVIL(s)} & $4.39$ & $\mathbf{69.7}$ & $\mathbf{9.89}$ & $\mathbf{121.3}$ & $11.91$ & $86.4$ & $11.46$ & $\mathbf{57.9}$ & $15.58$ & $92.3$ \\
\textbf{EVIL} & $\mathbf{4.05}$ & $71.0$ & $10.94$ & $166.1$ & $11.65$ & $84.4$ & $10.93$ & $71.2$ & $\mathbf{15.56}$ & $\mathbf{82.5}$ \\
\bottomrule
\end{tabular}
\vspace{-8pt}
\end{table}
Results for $N{=}10$ and $N{=}20$ follow similar trends and are reported in Appendix~\ref{app:tpp_tables}.

\begin{figure}[t]
\scriptsize
\begin{center}
\begin{minipage}[t]{0.32\columnwidth}
\captionsetup{type=table}
\caption{Runtime for one $N{=}5$ point-process evaluation. Other neural baselines are slower overall because they must also be trained per dataset.}
\label{tab:tpp-runtime}
\centering
\begin{tabular}{@{}lll@{}}
\toprule
Method & Hardware & Time \\
\midrule
FIM-PP(zs) & A100 40GB & 964s (16min) \\
\ours & 1 CPU core & 5s \\
\bottomrule
\end{tabular}
\end{minipage}\hfill
\begin{minipage}[t]{0.58\columnwidth}
\captionsetup{type=table}
\caption{Generalization to \dataset{MIMIC-II} \cite{lee2011open}, an unseen dataset with 75 marks (vs.\ max 22 during evolution). Baselines from \cite{panos2024decomposable}.}
\label{tab:mimic2}
\centering
\begin{tabular}{lcc}
\toprule
Method & RMSE & Accuracy \\
\midrule
THP \cite{zuo2021transformer} & 1.00 & 84.88 \\
SAHP \cite{zhang2020self} & 1.49 & 83.61 \\
A-NHP \cite{mei2022transformer} & 1.00 & 84.81 \\
IFTPP \cite{shchur2020intensity} & 0.74 & 85.13 \\
DTPP \cite{panos2024decomposable} & \textbf{0.72} & 85.51 \\
\midrule
\textbf{\ourssynth} & 0.90 & 83.14\\
\textbf{\ours} & 0.89 & \textbf{87.80}\\
\bottomrule
\end{tabular}
\end{minipage}
\end{center}
\vspace{-8pt}
\end{figure}

\begin{table}[t]
\caption{Next-event prediction ($N{=}1$) on two real-world datasets. Baseline results from \cite{cdiff,fim_pp}. Best in bold.}
\label{tab:next-event-pred}
\begin{center}
\scriptsize
\setlength{\tabcolsep}{4pt}
\begin{tabular}{l @{\hspace{10pt}} c c c @{\hspace{15pt}} c c c}
 & \multicolumn{3}{c}{\datasetbf{Taxi}} & \multicolumn{3}{c}{\datasetbf{Taobao}} \\
  \cmidrule(lr){2-4}  \cmidrule(lr){5-7}
Method & \RMSEdt & \accuracy & \SMAPEdt & \RMSEdt & \accuracy & \SMAPEdt \\
\midrule
\method{A-NHP}& $0.32$ & $\mathbf{0.91}$ & $85.13$ & $0.53$ & $0.47$ & $129.13$ \\
\method{Dual-TPP} & $0.34$ & $\mathbf{0.91}$ & $89.12$ & $0.53$ & $0.47$ & $131.43$ \\
\method{NHP}                    & $0.34$ & $\mathbf{0.91}$ & $90.63$ & $0.53$ & $0.46$ & $133.69$ \\
\method{IFTPP}   & $0.38$ & $0.90$ & $90.03$ & $0.53$ & $0.45$ & $126.01$ \\
\method{CDiff}                  & $0.34$ & $\mathbf{0.91}$ & $87.12$ & $0.52$ & $0.48$ & $127.12$ \\
\method{HawkesEM} & $0.34$ & $0.09$ & $79.93$ & $\mathbf{0.13}$ & $0.56$ & $111.57$ \\
\FIMzeroshot  & $0.34$ & $0.52$ & $98.03$ & $\mathbf{0.13}$ & $0.52$ & $112.89$ \\
\FIMfine  & $0.45$ & $\mathbf{0.91}$ & $85.91$ & $0.16$ & $0.60$ & $115.98$ \\
\midrule
\textbf{\ourssynth}      & $0.30$ & $\mathbf{0.91}$ & $\mathbf{69.49}$ & $\mathbf{0.13}$ & $\mathbf{0.61}$ & $\mathbf{95.71}$ \\
\textbf{\ours}      & $\mathbf{0.29}$ & $\mathbf{0.91}$ & $71.98$ & $\mathbf{0.13}$ & $0.59$ & $117.62$ \\
\end{tabular}

\end{center}
\vspace{-8pt}
\end{table}

\paragraph{Generalization.}
Table~\ref{tab:next-event-pred} shows next-event ($N{=}1$) prediction results on \dataset{Taxi} and \dataset{Taobao}.
On \dataset{Taxi}, both \ours variants match the best neural baselines in accuracy (0.91) while substantially improving sMAPE.
The \dataset{Taxi} dataset is special in the sense that it often contains alternations between marks (see Figure~\ref{fig:taxi_heuristic} in the appendix). This is not covered well by the synthetic training data of \cite{fim_pp}, which is why FIM-PP(zs) performs comparatively poorly.
Notably, \ourssynth however already achieves 91\% accuracy on \dataset{Taxi} using only synthetic training data, whereas the non-finetuned FIM-PP(zs) reaches only 52\%. This indicates that the LLM-evolved algorithm was able to generalize to special patterns that were not part of the synthetic training data.
\par
Moreover, Table~\ref{tab:mimic2} demonstrates the generalization capabilities of \ours on \dataset{MIMIC-II} \cite{lee2011open}, a medical event dataset with 75 mark types that was \emph{never seen during evolution} (the maximum during evolution was 22 marks since we evolved on the datasets of Table \ref{tab:long-horizon-5-compact} or synthetic data).
\ours achieves 87.8\% accuracy, surpassing all neural baselines that were specifically trained on this dataset.
This is particularly striking because neural TPP models typically have a fixed output dimension matching the training mark count and cannot easily accommodate new marks, whereas the \ours algorithm naturally handles arbitrary mark counts through its use of transition statistics computed at inference time.

\section{Markov Jump Processes}
\label{sec:mjp}

\subsection{Background}
A Markov jump process (MJP) \cite{norris1998markov} is a continuous-time stochastic process taking values in a finite state space $\{1, \dots, K\}$.
It is fully characterized by a rate matrix $Q \in \mathbb{R}^{K \times K}$ with non-negative off-diagonal entries and rows summing to zero, together with an initial distribution $\pi_0$.
The fundamental inference task is to estimate $Q$ and $\pi_0$ from noisy, discretely observed trajectories---a challenging inverse problem since exact jump times are unobserved.
Appendix~\ref{app:mjp_details} provides additional background on MJPs, the datasets, the evaluation metrics, and the DFR entropy-production calculation.

\subsection{Experimental Setup}
We evolve a function that takes discrete observation grids and state sequences and returns a rate matrix and initial distribution.
The initial program computes simple normalized transition counts.
The evolutions (100 iterations) took about 14 minutes wall-clock time.
Since ground-truth rate matrices are only available for synthetic data, \ourssynth is evolved using 7{,}200 synthetic MJPs from the same family as FIM-MJP \cite{fim_mjp}, with state spaces of size 2--6.

\paragraph{Baselines.}
We compare against NeuralMJP \cite{neural_mjp}, a neural network trained separately on each target dataset, and FIM-MJP \cite{fim_mjp}, a foundation model pretrained on 45K synthetic MJPs.
Both are substantially more complex architectures with millions of parameters.

\begin{table}[t]
\caption{DFR rate matrix parameters inferred by different methods, expressed as ratios to the ground truth (1.0 is perfect). $V$: potential strength, $r$/$b$: rate parameters.}
\label{tab:DFR}
\scriptsize
\begin{center}
\begin{tabular}{lccc}
    \toprule 
                 &    $V$ &   $r$ &  $b$ \\
    \midrule
    Ground Truth &  $1.00$ &   $1.00$ &   $1.00$ \\
    \midrule
    NeuralMJP  &  $1.06$ & $1.17$ & $1.14$ \\
    FIM-MJP  & $1.11$ & $\mathbf{0.99}$ & $\mathbf{0.98}$\\
    \textbf{\ourssynth} & $\mathbf{0.97}$ & $1.04$ & $1.06$\\
    \bottomrule
\end{tabular}
\end{center}
\vspace{-8pt}
\end{table}

\begin{table}[t]
    \caption{Time-averaged Hellinger distances ($\times 10^{-2}$, lower is better) between empirical and simulated processes. Mean (std.) over 100 histograms.}
    \label{tab:hellinger_distances}
\scriptsize
\begin{center}
    \begin{tabular}{lcccc}
    \toprule
    Model & \textsc{DFR} & \textsc{IonCh} & \textsc{ADP} & \textsc{PFold}\\
    \midrule
    NeuralMJP & $0.30\,(0.06)$ & $0.48\,(0.02)$ & $1.38\,(0.52)$ & $\mathbf{0.015\,(0.015)}$\\
    FIM-MJP & $0.27\,(0.06)$ & $0.41\,(0.02)$ & $1.39\,(0.47)$ & $\mathbf{0.014\,(0.014)}$\\
    \textbf{\ourssynth} & $\mathbf{0.16}\,(0.02)$ & $\mathbf{0.30}\,(0.06)$ & $\mathbf{0.93}\,(0.23)$ & $\mathbf{0.015\,(0.013)}$\\
    \bottomrule
    \end{tabular}
\end{center}
\vspace{-8pt}
\end{table}

\subsection{Discovered Algorithm}
The evolved MJP heuristic (see Pseudocode~\ref{algpseud:mjp} and Python Algorithm~\ref{alg:mjp_synthetic} in Appendix~\ref{app:algorithms}) is:
\emph{(1) estimate the initial distribution from smoothed counts of the first observations; (2) for each state, accumulate how long trajectories stay there and how often they leave; (3) set the exit rate to approximately exits divided by exposure time, with simple clipping and noise filtering; (4) distribute this exit rate across destination states using recency-weighted transition counts with smoothing.}

\subsection{Results and Analysis}
We begin with the Discrete Flashing Ratchet (DFR) \cite{ajdari1992mouvement}, a structured 6-state non-equilibrium system that serves as a controlled benchmark for MJP inference: it is simple enough that its generator is determined by a few physically meaningful parameters, but rich enough to induce derived observables such as entropy production.
We therefore report recovery of the underlying parameters $(V,r,b)$ rather than listing the entire rate matrix entry-by-entry. Appendix~\ref{app:mjp_details} gives the exact DFR formulas and more background.
Table~\ref{tab:DFR} shows that \ourssynth follows the overall DFR structure well and is competitive with FIM-MJP, with the clearest advantage on the potential strength $V$ that controls the asymmetry of the ratchet.

\begin{figure}[t]
    \centering
    \includegraphics[width=0.5\linewidth]{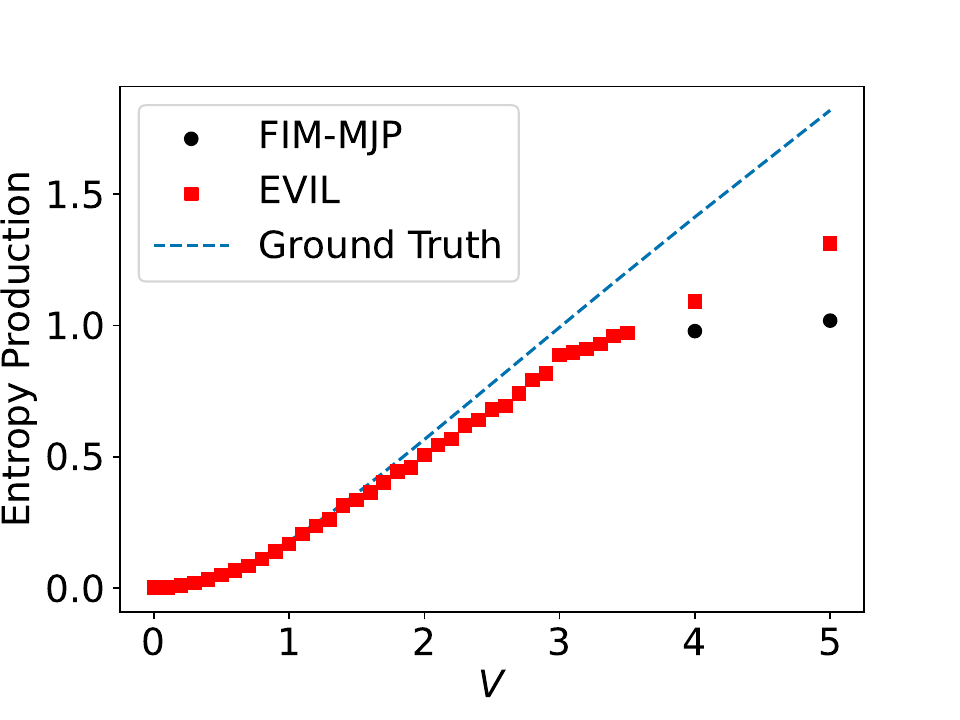}
    \caption{Entropy production on the DFR. This serves as a stress test for when inference becomes harder as the rates become increasingly dissimilar. Across the shown potentials, \ourssynth tracks the ground truth more closely than FIM-MJP.}
    \label{fig:dfr_entropy}
    \vspace{-8pt}
\end{figure}

Figure~\ref{fig:dfr_entropy} then evaluates a more demanding downstream quantity on the same benchmark: entropy production, which is nonzero in the DFR because the ratchet sustains irreversible probability currents.
The qualitative picture is again favorable to \ourssynth, which stays closer to the ground truth than FIM-MJP across the shown range.
At high voltages the DFR becomes intrinsically harder because a few transitions dominate while others become extremely rare: some rates differ by roughly four orders of magnitude, so the rare transitions are only weakly represented in the observed paths.
That makes entropy production sensitive to estimation noise in those low-count entries, but \ourssynth still recovers the very rarely observed transitions more precisely than FIM-MJP.

Table~\ref{tab:hellinger_distances} then broadens the picture beyond DFR and asks whether the inferred generators produce realistic samples on several datasets, including real-world phenomena such as ion-channel recordings \cite{gazzarrini2006chlorella} and molecular dynamics systems \cite{mironov19} where accurate MJP inference is scientifically useful because the recovered generator is used to characterize latent kinetics rather than just to make one-step predictions.
Here we use the Hellinger distance \cite{hellinger1909neue} as a similarity measure between the predicted path distribution and the true path distribution; this lets us assess sampling accuracy even when the ground-truth rate matrices are unavailable, as is the case for the real-world datasets. Appendix~\ref{app:mjp_details} briefly explains the metric and its use.
Interestingly, while NeuralMJP and FIM-MJP achieve very similar Hellinger distances to each other, \ourssynth is consistently better on DFR, IonCh, and ADP.

\section{Time Series Imputation}
\label{sec:imputation}

\subsection{Background}
Time series imputation aims to reconstruct missing values in partially observed multivariate time series.
In many applications, these observed sequences are not arbitrary collections of numbers, but partial observations of an underlying trajectory generated by a dynamical system.
From this perspective, both observed and missing values lie on a latent time-evolving process, so imputation can be viewed as inferring the trajectory that best explains the available observations and then reading off the unobserved values.
We consider two settings: point-wise missing patterns (random individual values are missing) and window-based missing patterns (contiguous blocks of observations are absent).

\subsection{Experimental Setup}
We evolve a function that takes observation values (with NaN at missing positions), timestamps, and a prediction mask, and returns fully imputed values.
The initial program uses simple last-value prediction.
The evolutions use 6 datasets: 4 with 50\% point-wise missing patterns (Beijing, Italy, GuangZhou, PeMS) and 2 Motion Capture variants with 20\% windowed missing patterns. The remaining 4 point-wise datasets (Pedestrian, Solar, ETT\_h1, Electricity) are held out entirely and used only for zero-shot generalization evaluation. Appendix~\ref{app:imputation_details} summarizes these datasets and their sources.
The evolutions (100 iterations) took about 37 minutes wall-clock time.

\paragraph{Baselines.}
We compare against BRITS \cite{cao2018brits} (bidirectional recurrent), SAITS \cite{du2023saits} (self-attention), GP-VAE \cite{fortuin2020gpvae} (Gaussian process VAE), CSDI \cite{tashiro2021csdi} (conditional score-based diffusion), BayOTIDE \cite{fang2023bayotide} (Bayesian decomposition), LatentODE \cite{rubanova2019latent} (neural ODE), and FIM-$\ell$ \cite{seifner2025zeroshot} (foundation model pretrained on 2M+ ODE solutions).
All baselines except FIM-$\ell$ require per-dataset training.
For the Motion Capture window-imputation setting, we additionally report cubic spline interpolation and a Savitzky--Golay filtered spline baseline (Spline(F)) as included in \cite{seifner2025zeroshot}.

\begin{table}[t]
\caption{MAE on 8 datasets with 50\% point-wise missing. \ours was evolved on the first four datasets (Beij., Italy, GZ, PeMS); the remaining four (Ped., Solar, ETT, Elec.) are zero-shot generalization targets. Baselines from \cite{fang2023bayotide,du2024tsi,seifner2025zeroshot}. Best in bold.}
\label{tab:imputation_pointwise}
\scriptsize
\setlength{\tabcolsep}{3pt}
\begin{center}
    \begin{tabular}{lcccccccc}
    \toprule
    & \multicolumn{2}{c}{\textit{Air quality}} & \multicolumn{2}{c}{\textit{Traffic}} & & \multicolumn{3}{c}{\textit{Electricity}} \\
    \cmidrule(lr){2-3}\cmidrule(lr){4-5}\cmidrule(lr){7-9}
    Method & Beij. & Italy & GZ & PeMS & Ped. & Solar & ETT & Elec. \\
    \midrule
    BRITS & .169 & .321 & 3.34 & \textbf{.287} & .259 & 1.99 & .238 & 1.12 \\
    SAITS & .194 & .285 & 3.39 & .302 & \textbf{.205} & 1.83 & \textbf{.223} & 1.40 \\
    GP-VAE & .258 & .453 & 3.42 & .346 & .451 & 1.81 & .414 & 1.10 \\
    CSDI & \textbf{.144} & .958 & 3.20 & .288 & .351 & .804 & .318 & .798 \\
    BayOTIDE & -- & -- & 2.69 & -- & -- & .734 & -- & -- \\
    FIM-$\ell$ & .166 & .215 & 2.43 & .365 & .273 & .595 & .279 & .083 \\
    \midrule
    \textbf{\ours} & .152 & \textbf{.204} & \textbf{2.09} & .338 & .242 & \textbf{.413} & .267 & \textbf{.078} \\
    \bottomrule
    \end{tabular}
\end{center}
\vspace{-8pt}
\end{table}

\begin{figure*}[t]
    \centering
    \begin{minipage}[t]{0.38\textwidth}
        \vspace{0pt}
        \captionsetup{type=table}
        \caption{MAE on Motion Capture (20\% windowed missing). Baselines from \cite{seifner2025zeroshot}.}
        \label{tab:imputation_mocap}
        \scriptsize
        \setlength{\tabcolsep}{2pt}
        \centering
        \begin{tabular}{lcc}
            \toprule
            Model & MC (PCA) & MC (No PCA) \\
            \midrule
            LatentODE & $\mathbf{1.66 \pm 0.99}$ & -- \\
            Spline & $3.36 \pm 1.18$ & $4.21 \pm 1.44$ \\
            Spline(F) & $2.90 \pm 0.87$ & $3.00 \pm 0.88$ \\
            FIM-$\ell$ & $1.77 \pm 0.63$ & $\mathbf{1.61 \pm 0.45}$ \\
            \midrule
            \textbf{\ours} & $2.98 \pm 2.04$ & $3.34 \pm 2.07$ \\
            \bottomrule
        \end{tabular}
    \end{minipage}\hfill
    \begin{minipage}[t]{0.48\textwidth}
        \vspace{0pt}
        \centering
        \includegraphics[width=0.98\linewidth]{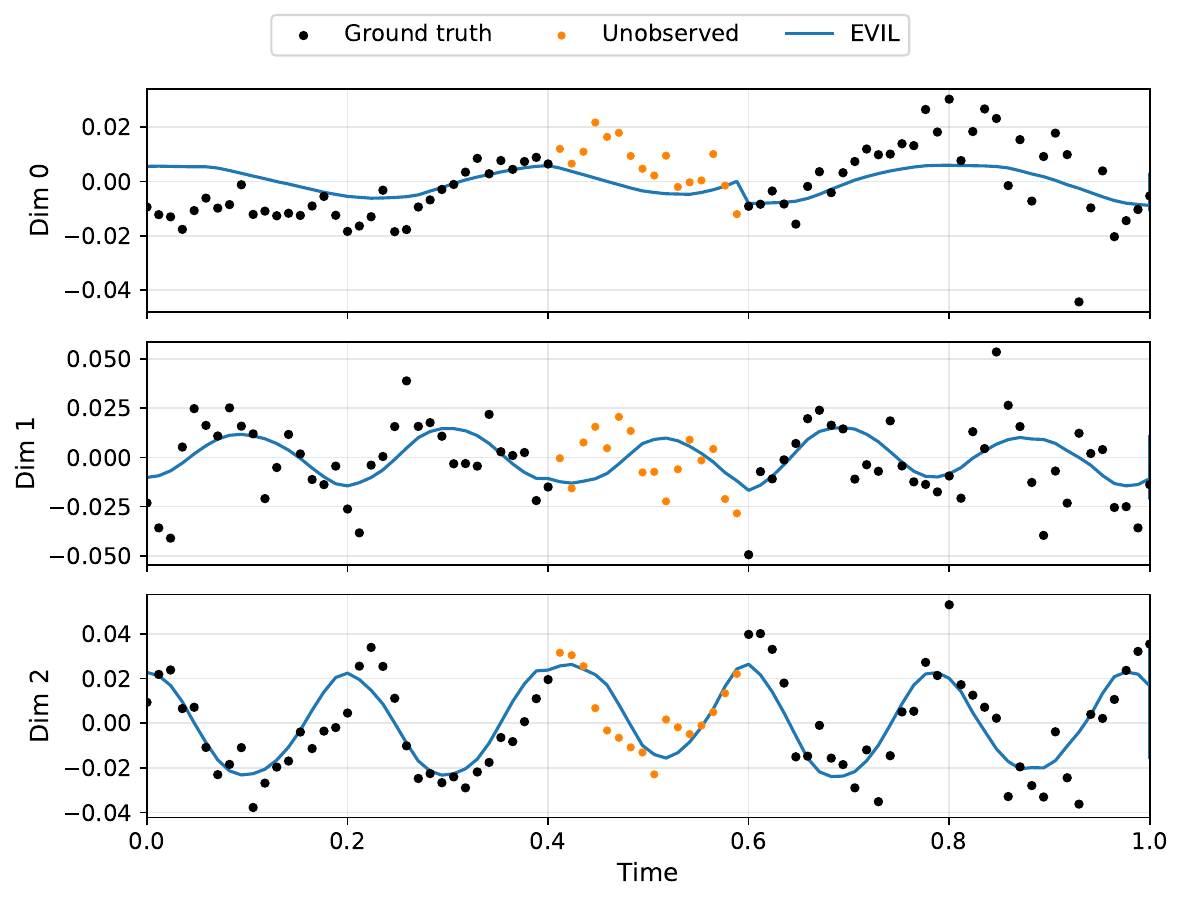}
        \caption{Imputation on Motion Capture. The motif retrieval strategy enables \ours to predict non-trivial patterns in missing windows.}
        \label{fig:mocap}
    \end{minipage}
    \vspace{-8pt}
\end{figure*}

\subsection{Discovered Algorithm}
The evolved imputation heuristic (see Pseudocode~\ref{algpseud:imputation} and Python Algorithm~\ref{alg:imputation} in Appendix~\ref{app:algorithms}) is:
\emph{(1) detect contiguous missing blocks; (2) if a gap is long, perform motif retrieval by searching earlier in the same series for a recurring local pattern whose preceding context matches the context before the gap, then copy the following pattern with a level shift; (3) otherwise, fill the gap by time-aware linear interpolation.}

\subsection{Results and Analysis}
Table~\ref{tab:imputation_pointwise} shows point-wise imputation results.
The simple program found by \ours performs competitively on all 8 datasets, including the 4 held-out datasets (Ped., Solar, ETT, Elec.) that were never seen during evolution, demonstrating zero-shot generalization.
This is consistent with findings from TSI-Bench \cite{du2024tsi}, which reported that simple interpolation methods often perform surprisingly well on point-wise missing patterns.

Table~\ref{tab:imputation_mocap} shows windowed imputation on Motion Capture.
Here, FIM-$\ell$ outperforms \ours, suggesting that reconstructing contiguous missing windows benefits from the learned dynamical priors that FIM-$\ell$ acquires during pretraining on ODE solutions.
At the same time, the variances are high across all methods, so this benchmark might be less conclusive.
Still, \ours discovers a simple nonlinear strategy based on \emph{motif} retrieval \cite{Lin2002FindingMI}: it searches earlier in the same sequence for a segment whose recent context resembles the context before the missing window, and then reuses the continuation that followed that earlier segment.
This already allows \ours to reconstruct structured oscillations in some cases (Figure~\ref{fig:mocap}).
Interestingly, the concurrent \emph{context parroting} baseline of \citet{zhang2026context} also relies on a motif-based strategy, suggesting that motif retrieval may be a naturally strong inductive bias for time series imputation that both human-designed and LLM-discovered methods converge on independently.\footnote{The context parroting paper has been released after the knowledge cutoff of the LLM, so the LLM seemingly discovered it independently.}

\section{Conclusion}
We explored whether LLM-guided evolutionary search can discover explicit, interpretable algorithms for dynamical systems inference.
Across three distinct tasks, the search successfully identified compact Python functions that perform zero-shot inference without per-dataset training.
While they do not always match deep neural networks on tasks requiring rich learned representations (such as windowed time series imputation) these heuristics are surprisingly competitive on standard benchmarks.
At the same time, they evaluate orders of magnitude faster, require minimal compute to discover, and remain fully transparent.

These results demonstrate that complicated neural architectures might not always be necessary to achieve strong predictive performance in dynamical systems.
Going forward, we hope \ours serves as a grounded, computationally cheap baseline to verify when the added complexity of a new model is actually justified, offering a practical step toward more transparent scientific computing.

\section{Limitations}
The main limitation of \ours is its weaker performance on tasks that heavily benefit from dataset-specific learned representations, such as windowed time series imputation (see Section~\ref{sec:imputation}).
Assessing novelty of the discovered algorithms is also difficult: while they seem intuitive in hindsight, they combine known concepts with optimizations (e.g., recency weightings, smoothing and clipping heuristics) that might be tedious for humans to design; though closely related approaches may already be known to domain experts.
Finally, the discovered algorithms are deterministic; exploring whether LLM-evolution can discover stochastic, uncertainty-aware algorithms remains an open direction.

\begin{ack}
The author would like to thank Patrick Seifner, Manuel Hinz, Injamam Karim, Ramses Sanchez, and Armin Berger for useful discussions.

This research has been funded by the Federal Ministry of Education and Research of Germany and the state of North-Rhine Westphalia as part of the Lamarr Institute for Machine Learning and Artificial Intelligence.

\end{ack}

\bibliographystyle{plainnat}
\bibliography{bib}


\appendix

\section{Additional Temporal Point Process Details}
\label{app:tpp_details}

\subsection{Mathematical Setup}

A marked temporal point process (MTPP) is a random sequence of events
\(
\{(t_i, k_i)\}_{i=1}^n
\)
with strictly increasing times $t_i \in \mathbb{R}_+$ and discrete marks $k_i \in \{1, \dots, K\}$ \cite{daley2003introduction}. Writing the history before time $t$ as
\(
\mathcal{H}_t = \{(t_i, k_i) : t_i < t\},
\)
the standard object is the mark-specific conditional intensity $\lambda_k(t \mid \mathcal{H}_t)$, whose value can be interpreted as the instantaneous rate at which an event of type $k$ occurs at time $t$ given the past. Summing over marks gives the total intensity,
\(
\lambda(t \mid \mathcal{H}_t) = \sum_{k=1}^K \lambda_k(t \mid \mathcal{H}_t),
\)
and the corresponding sequence density on an interval $[0,T]$ can be written as
\begin{equation}
p\!\left(\{(t_i, k_i)\}_{i=1}^n\right)
=
\left[
\prod_{i=1}^n \lambda_{k_i}(t_i \mid \mathcal{H}_{t_i})
\right]
\exp\!\left(
-\int_0^T \lambda(s \mid \mathcal{H}_s)\, ds
\right).
\end{equation}
This formalism makes clear that prediction has two coupled parts: forecasting \emph{when} the next event happens and \emph{which mark} it carries. Classical Hawkes models specify these intensities through hand-designed self- and cross-excitation kernels \cite{hawkes1971spectra}, whereas recent neural approaches learn flexible history representations from data. In our setting, the inference problem is in-context next-event or long-horizon forecasting: given a small set of reference sequences from one system and a partial history from a test sequence, the algorithm predicts the next $N$ event times and marks without per-dataset retraining.

\subsection{Datasets}

For evolution, we use 33 synthetic point-process datasets generated from the same broad synthetic distribution used in FIM-PP \cite{fim_pp}. In particular, we reuse their point-process family and synthetic data-generation setup, which spans Poisson, Hawkes, and periodic regimes with randomized time-dependent base intensities, mark-to-mark interaction kernels, and excitatory, inhibitory, or neutral couplings between marks; our subsampled training collection contains between 1 and 22 marks and 25 sampled sequences per dataset. These synthetic tasks are meant to expose the search to qualitatively different dynamics, ranging from nearly memoryless arrivals to strongly history-dependent excitation patterns. The full \ours variant additionally incorporates small subsamples from five real datasets, while \ourssynth uses only the synthetic collection.

\paragraph{\dataset{Taxi}.}
This dataset is derived from New York City taxi trip logs, using the benchmark preprocessing adopted in prior TPP work \cite{cdiff}. Each sequence corresponds to one driver's activity, and each event records either a pick-up or a drop-off together with the borough in which it occurred, yielding 10 marks in total. We follow the standard benchmark subset of 2,000 drivers.

\begin{figure}[t]
    \centering
    \includegraphics[width=\linewidth]{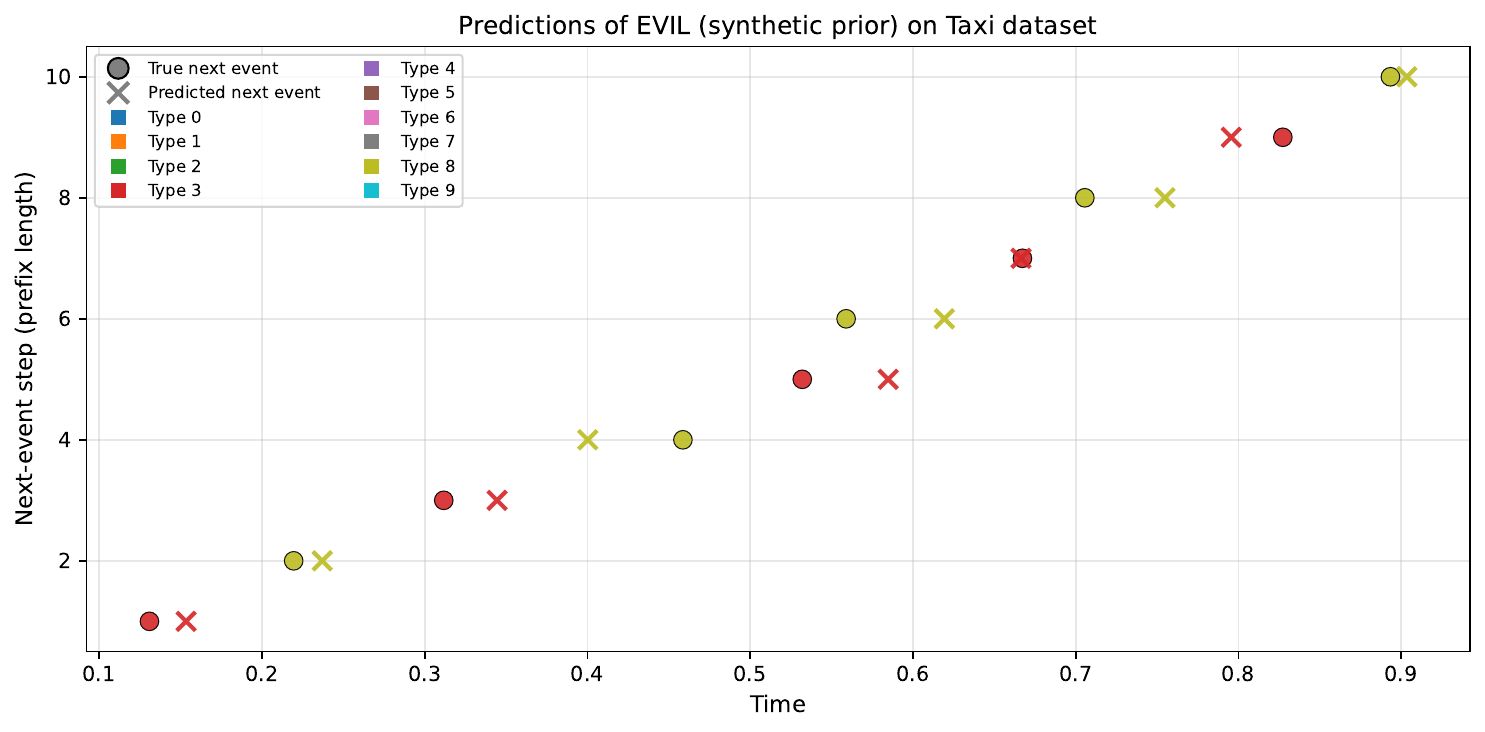}
    \caption{Illustration of mark alternation in \dataset{Taxi} and how the evolved next-event heuristic handles it. The dataset often alternates between marks (e.g., pick-up vs.\ drop-off); this pattern is not covered by FIM-PP's synthetic training data, whereas \ourssynth generalizes to it from synthetic evolution.}
    \label{fig:taxi_heuristic}
    \vspace{-8pt}
\end{figure}

\paragraph{\dataset{Taobao}.}
This dataset comes from user interaction logs on the Taobao e-commerce platform \cite{Zhu_2018}. Each sequence tracks one anonymous user's actions over time, and the 17 marks correspond to different interaction or product-category groups. We use the benchmark subset of 2,000 active users.

\paragraph{\dataset{StackOverflow}.}
Here each sequence records the times at which a user receives badges on StackOverflow \cite{snapnets}. The marks are the 22 badge types, and the benchmark subset contains 2,200 active users.

\paragraph{\dataset{Amazon}.}
This dataset consists of user review histories on Amazon over a multi-year period \cite{ni-etal-2019-justifying}. Each event is a review timestamp, and the mark is the corresponding product category; the benchmark version uses 16 marks and 5,200 active users.

\paragraph{\dataset{Retweet}.}
This dataset represents retweet activity over time \cite{pmlr-v28-zhou13}. Each sequence is a user's retweet history, and the three marks coarse-grain the influence of the original poster into small, medium, and large accounts. We use the benchmark subset of 5,200 active users.

\paragraph{\dataset{MIMIC-II}.}
For out-of-distribution evaluation, we additionally report transfer to \dataset{MIMIC-II}, a medical event dataset derived from intensive-care records \cite{lee2011open}. In our benchmark it contains 75 marks, far beyond the maximum of 22 marks seen during evolution, so it provides a clean test of whether the evolved heuristic scales to much richer event vocabularies without retraining.

For all five real-world benchmark datasets, we follow the preprocessing and train/test splits used by prior TPP work, especially CDiff and FIM-PP \cite{cdiff,fim_pp}.

\subsection{Evaluation Metrics}
\label{app:eval_metrics}

Following the benchmark protocol of CDiff and FIM-PP \cite{cdiff,fim_pp}, we evaluate predicted future sequences
\(
\hat{\mathbf{s}}_{\mathrm{future}} = \{(\hat{t}_i, \hat{k}_i)\}_{i=1}^N
\)
against ground truth
\(
\mathbf{s}_{\mathrm{future}} = \{(t_i, k_i)\}_{i=1}^N.
\)
For next-event prediction, we report mark accuracy together with an error metric on the predicted inter-event time. For long-horizon forecasting, we report Optimal Transport Distance (OTD) \cite{mei-2019-imputing-missing-otd}, which measures the overall discrepancy between predicted and true event sequences through edit-style costs on timing, marks, and event counts.

Let $\Delta t_i = t_i - t_{i-1}$ and let $C_{j,k}$ and $\hat{C}_{j,k}$ denote the true and predicted counts of mark $k$ in test sequence $j$. Then the main quantitative metrics are
\begin{equation}
\mathrm{Acc}
=
\frac{1}{m}\sum_{j=1}^m \frac{1}{N}\sum_{i=1}^N \mathbb{I}(k_{j,i} = \hat{k}_{j,i}),
\end{equation}
\begin{equation}
\mathrm{RMSE}_e
=
\sqrt{\frac{1}{m}\sum_{j=1}^m \sum_{k \in \mathcal{K}} (C_{j,k} - \hat{C}_{j,k})^2},
\end{equation}
\begin{equation}
\mathrm{RMSE}_{\Delta t}
=
\sqrt{\frac{1}{m}\sum_{j=1}^m \frac{1}{N}\sum_{i=1}^N (\Delta t_{j,i} - \widehat{\Delta t}_{j,i})^2},
\end{equation}
and
\begin{equation}
\mathrm{sMAPE}_{\Delta t}
=
\frac{100}{m}\sum_{j=1}^m \frac{1}{N}\sum_{i=1}^N
\frac{2|\Delta t_{j,i} - \widehat{\Delta t}_{j,i}|}{|\Delta t_{j,i}| + |\widehat{\Delta t}_{j,i}|}.
\end{equation}
Here, accuracy is higher-is-better, while OTD, $\mathrm{RMSE}_e$, $\mathrm{RMSE}_{\Delta t}$, and $\mathrm{sMAPE}_{\Delta t}$ are lower-is-better. Intuitively, $\mathrm{Acc}$ checks whether the predicted marks are correct at the right positions, $\mathrm{RMSE}_e$ compares the overall mark histogram over the forecast horizon, and the last two metrics measure the quality of the predicted inter-event times.

\subsection{Baseline Models}

NHP and A-NHP \cite{mei2017neural,mei2022transformer} are neural intensity models, IFTPP \cite{shchur2020intensity} is an intensity-free neural alternative, HawkesEM \cite{hawkesem} is a classical expectation maximization baseline implemented via \texttt{tick} \cite{tick}, Dual-TPP and HYPRO \cite{deshpande2021long,xue2022hypro} are long-horizon forecasting architectures, and TCDDM together with CDiff \cite{lin2022exploring,cdiff} are generative diffusion-style approaches for event-sequence prediction. We also compare to FIM-PP \cite{fim_pp}, a pretrained transformer-based foundation inference model that is trained on a large synthetic corpus of point processes and then applied either zero-shot or after lightweight finetuning to a target dataset. All of these baselines are substantially more complex than our final heuristic, and except for the pretrained FIM-PP variants they require separate training for each dataset.

\section{Long-Horizon Temporal Point Process Tables}
\label{app:tpp_tables}

\begin{table}[h]
\caption{
Prediction of $N=20$ events in test sequences of five real-world datasets. 
Error-bars indicate the standard deviation over $10$ trials. 
Results for the baseline methods were extracted from \cite{cdiff} and \cite{fim_pp}. Best results are bold. 
}
\label{tab:long-horizon-pred-20}
\small
\begin{center}


\begin{tabular}{llrrrr}
Dataset              & Method   & \OTD                                   & \RMSEe                           & \RMSEdt                         & \SMAPEdt                                 \\
\midrule
\multirow{10}{*}{\datasetbf{Taxi}} & \method{HYPRO}    & $21.653 {\tinymath{\pm 0.163}}      $ & $1.231 {\tinymath{\pm 0.015}}$ & $0.372 {\tinymath{\pm 0.004}}$ & $93.803 {\tinymath{\pm 0.454}}$ \\
                       & \method{Dual-TPP} & $24.483 {\tinymath{\pm 0.383}}$ & $1.353 {\tinymath{\pm 0.037}}$ & $0.402 {\tinymath{\pm 0.006}}$ & $95.211 {\tinymath{\pm 0.187}}$ \\
                       & \method{A-NHP}    & $24.762 {\tinymath{\pm 0.217}}$ & $1.276 {\tinymath{\pm 0.015}}$ & $0.430 {\tinymath{\pm 0.003}}$ & $97.388 {\tinymath{\pm 0.381}}$ \\
                       & \method{NHP}      & $25.114 {\tinymath{\pm 0.268}}$ & $1.297 {\tinymath{\pm 0.019}}$ & $0.399 {\tinymath{\pm 0.040}}$ & $96.459 {\tinymath{\pm 0.521}}$ \\
                       & \method{IFTPP}    & $24.053 {\tinymath{\pm 0.609}}$ & $1.364 {\tinymath{\pm 0.032}}$ & $0.384 {\tinymath{\pm 0.005}}$ & $95.719 {\tinymath{\pm 0.779}}$ \\
                       & \method{TCDDM}    & $22.148 {\tinymath{\pm 0.529}}$ & $1.309 {\tinymath{\pm 0.030}}$ & $0.382 {\tinymath{\pm 0.019}}$ & $90.596 {\tinymath{\pm 0.574}}$ \\
                       & \method{CDiff}    & $21.013 {\tinymath{\pm 0.158}}$ & $1.131 {\tinymath{\pm 0.017}}$ & $0.351 {\tinymath{\pm 0.004}}$ & $87.993 {\tinymath{\pm 0.178}}$ \\
                       & \FIMzeroshot  & $23.145$ \tinymath{\pm 0.073} & $1.421$ \tinymath{\pm 0.014} & $0.277$ \tinymath{\pm 0.000} & $76.765$ \tinymath{\pm 0.386} \\
   
                        &\FIMfine  & $17.914$ \tinymath{\pm 0.117} & $\mathbf{0.705}$ \tinymath{\pm 0.006} & $0.314$ \tinymath{\pm 0.004} & $76.828$ \tinymath{\pm 0.549}\\

                        & \textbf{\ourssynth} & $20.007$ \tinymath{\pm 0.000} & $0.853$ \tinymath{\pm 0.000} & $0.272$ \tinymath{\pm 0.000} & $\mathbf{70.750}$ \tinymath{\pm 0.000} \\

                        & \textbf{\ours} & $\mathbf{16.969}$ \tinymath{\pm 0.000} & $0.725$ \tinymath{\pm 0.000} & $\mathbf{0.264}$ \tinymath{\pm 0.000} & $72.324$ \tinymath{\pm 0.000} \\

\midrule
\multirow{10}{*}{\datasetbf{Taobao}} & \method{HYPRO}    & $\mathbf{44.336} {\tinymath{\pm 0.127}}      $ & $2.710 {\tinymath{\pm 0.021}}$ & $0.594 {\tinymath{\pm 0.030}}$ & $134.922 {\tinymath{\pm 0.473}}$ \\
                         & \method{Dual-TPP} & $47.324 {\tinymath{\pm 0.541}}$ & $3.237 {\tinymath{\pm 0.049}}$ & $0.871 {\tinymath{\pm 0.005}}$ & $141.687 {\tinymath{\pm 0.431}}$ \\
                         & \method{A-NHP}    & $45.555 {\tinymath{\pm 0.345}}$ & $2.737 {\tinymath{\pm 0.021}}$ & $0.708 {\tinymath{\pm 0.010}}$ & $134.582 {\tinymath{\pm 0.920}}$ \\
                         & \method{NHP}      & $48.131 {\tinymath{\pm 0.297}}$ & $3.355 {\tinymath{\pm 0.030}}$ & $0.837 {\tinymath{\pm 0.009}}$ & $137.644 {\tinymath{\pm 0.764}}$ \\
                         & \method{IFTPP}    & $45.757 {\tinymath{\pm 0.287}}$ & $3.193 {\tinymath{\pm 0.043}}$ & $0.575 {\tinymath{\pm 0.012}}$ & $127.436 {\tinymath{\pm 0.606}}$ \\
                         & \method{TCDDM}    & $45.563 {\tinymath{\pm 0.889}}$ & $2.850 {\tinymath{\pm 0.058}}$ & $0.569 {\tinymath{\pm 0.015}}$ & $126.512 {\tinymath{\pm 0.491}}$ \\
                         & \method{CDiff}    & $44.621 {\tinymath{\pm 0.139}}$ & $2.653 {\tinymath{\pm 0.022}}$ & $\mathbf{0.551} {\tinymath{\pm 0.002}}$ & $125.685 {\tinymath{\pm 0.151}}$ \\
                         & \FIMzeroshot  & $64.281$ \tinymath{\pm 0.077} & $3.949$ \tinymath{\pm 0.010} & $1.988$ \tinymath{\pm 0.006} & $169.687$ \tinymath{\pm 0.089}\\

                         & \FIMfine  &  $60.106$ \tinymath{\pm 0.464} & $\mathbf{2.428}$ \tinymath{\pm 0.005} & $16.068$ \tinymath{\pm 0.109} & $152.528$ \tinymath{\pm 0.377} \\

                         & \textbf{\ourssynth} & $49.979$ \tinymath{\pm 0.000} & $3.227$ \tinymath{\pm 0.000} & $1.558$ \tinymath{\pm 0.000} & $\mathbf{119.260}$ \tinymath{\pm 0.000} \\

                         & \textbf{\ours} & $52.356$ \tinymath{\pm 0.000} & $2.979$ \tinymath{\pm 0.000} & $1.688$ \tinymath{\pm 0.000} & $170.464$ \tinymath{\pm 0.000} \\

\midrule
\multirow{10}{*}{\datasetbf{StackOverflow}} & \method{HYPRO}    & $42.359 {\tinymath{\pm 0.170}}      $ & $1.140 {\tinymath{\pm 0.014}}      $ & $1.554 {\tinymath{\pm 0.010}}$ & $110.988 {\tinymath{\pm 0.559}}$ \\
                                & \method{Dual-TPP} & $41.752 {\tinymath{\pm 0.200}}$ & $\mathbf{1.134} {\tinymath{\pm 0.019}}$ & $1.514 {\tinymath{\pm 0.017}}$ & $117.582 {\tinymath{\pm 0.420}}$ \\
                                & \method{A-NHP}    & $42.591 {\tinymath{\pm 0.408}}$ & $1.142 {\tinymath{\pm 0.011}}$ & $1.340 {\tinymath{\pm 0.006}}$ & $108.542 {\tinymath{\pm 0.531}}$ \\
                                & \method{NHP}      & $43.791 {\tinymath{\pm 0.147}}$ & $1.244 {\tinymath{\pm 0.030}}$ & $1.487 {\tinymath{\pm 0.004}}$ & $116.952 {\tinymath{\pm 0.404}}$ \\
                                & \method{IFTPP}    & $46.280 {\tinymath{\pm 0.892}}$ & $1.447 {\tinymath{\pm 0.057}}$ & $1.669 {\tinymath{\pm 0.005}}$ & $115.122 {\tinymath{\pm 0.627}}$ \\
                                & \method{TCDDM}    & $42.128 {\tinymath{\pm 0.591}}$ & $1.467 {\tinymath{\pm 0.014}}$ & $1.315 {\tinymath{\pm 0.004}}$ & $107.659 {\tinymath{\pm 0.934}}$ \\
                                & \method{CDiff}    & $41.245 {\tinymath{\pm 1.400}}$ & $1.141 {\tinymath{\pm 0.007}}$ & $1.199 {\tinymath{\pm 0.006}}$ & $106.175 {\tinymath{\pm 0.340}}$ \\
                                &\FIMzeroshot  & $49.259$ \tinymath{\pm 0.056} & $2.393$ \tinymath{\pm 0.015} & $1.068$ \tinymath{\pm 0.002} & $96.364$ \tinymath{\pm 0.048} \\
              
                                & \FIMfine & $\mathbf{39.792}$ \tinymath{\pm 0.042} & $1.336$ \tinymath{\pm 0.030} & $1.018$ \tinymath{\pm 0.003} & $88.248$ \tinymath{\pm 0.189} \\

                                & \textbf{\ourssynth} & $49.396$ \tinymath{\pm 0.000} & $2.858$ \tinymath{\pm 0.000} & $1.010$ \tinymath{\pm 0.000} & $88.734$ \tinymath{\pm 0.000} \\
                                
                                & \textbf{\ours} & $46.141$ \tinymath{\pm 0.000} & $2.839$ \tinymath{\pm 0.000} & $\mathbf{0.977}$ \tinymath{\pm 0.000} & $\mathbf{87.773}$ \tinymath{\pm 0.000} \\

\midrule
\multirow{10}{*}{\datasetbf{Amazon}}        & \method{HYPRO}    & $38.613 {\tinymath{\pm 0.536}}$ & $\mathbf{2.007} {\tinymath{\pm 0.054}}      $ & $0.477 {\tinymath{\pm 0.010}}$ & $82.506 {\tinymath{\pm 0.840}}      $ \\
                                & \method{Dual-TPP} & $42.646 {\tinymath{\pm 0.752}}$ & $2.562 {\tinymath{\pm 0.202}}$ & $0.482 {\tinymath{\pm 0.012}}$ & $86.453 {\tinymath{\pm 2.044}}$ \\
                                & \method{A-NHP}    & $39.480 {\tinymath{\pm 0.326}}$ & $2.166 {\tinymath{\pm 0.026}}$ & $0.476 {\tinymath{\pm 0.033}}$ & $84.323 {\tinymath{\pm 1.815}}$ \\
                                & \method{NHP}      & $42.571 {\tinymath{\pm 0.293}}$ & $2.561 {\tinymath{\pm 0.060}}$ & $0.519 {\tinymath{\pm 0.023}}$ & $92.053 {\tinymath{\pm 1.553}}$ \\
                                & \method{IFTPP}    & $43.820 {\tinymath{\pm 0.232}}$ & $3.050 {\tinymath{\pm 0.286}}$ & $0.481 {\tinymath{\pm 0.145}}$ & $90.910 {\tinymath{\pm 1.611}}$ \\
                                & \method{TCDDM}    & $42.245 {\tinymath{\pm 0.174}}$ & $2.998 {\tinymath{\pm 0.115}}$ & $0.476 {\tinymath{\pm 0.111}}$ & $83.826 {\tinymath{\pm 1.508}}$ \\
                                & \method{CDiff}    & $37.728 {\tinymath{\pm 0.199}}$ & $2.091 {\tinymath{\pm 0.163}}$ & $0.464 {\tinymath{\pm 0.086}}$ & $81.987 {\tinymath{\pm 1.905}}$ \\
                                & \FIMzeroshot  & $46.219$ \tinymath{\pm 0.108} & $2.073$ \tinymath{\pm 0.012} & $0.464$ \tinymath{\pm 0.001} & $128.635$ \tinymath{\pm 0.398}\\
     
                                & \FIMfine & $\mathbf{37.208}$ \tinymath{\pm 0.098} & $\mathbf{2.030}$ \tinymath{\pm 0.019} & $0.366$ \tinymath{\pm 0.001} & $81.188$ \tinymath{\pm 0.142} \\

                                & \textbf{\ourssynth} & $46.423$ \tinymath{\pm 0.000} & $3.738$ \tinymath{\pm 0.000} & $0.358$ \tinymath{\pm 0.000} & $\mathbf{61.841}$ \tinymath{\pm 0.000} \\
                                
                                & \textbf{\ours} & $43.932$ \tinymath{\pm 0.000} & $3.277$ \tinymath{\pm 0.000} & $\mathbf{0.326}$ \tinymath{\pm 0.000} & $78.108$ \tinymath{\pm 0.000} \\

\midrule
\multirow{10}{*}{\datasetbf{Retweet}}       & \method{HYPRO}    & $61.031 {\tinymath{\pm 0.092}}$ & $2.623 {\tinymath{\pm 0.036}}$ & $30.100 {\tinymath{\pm 0.413}}$ & $106.110 {\tinymath{\pm 1.505}}      $ \\
                                & \method{Dual-TPP} & $61.095 {\tinymath{\pm 0.101}}$ & $2.679 {\tinymath{\pm 0.026}}$ & $28.914 {\tinymath{\pm 0.300}}$ & $106.900 {\tinymath{\pm 1.293}}$ \\
                                & \method{A-NHP}    & $60.634 {\tinymath{\pm 0.097}}$ & $2.561 {\tinymath{\pm 0.054}}$ & $28.812 {\tinymath{\pm 0.272}}$ & $107.234 {\tinymath{\pm 1.293}}$ \\
                                & \method{NHP}      & $60.953 {\tinymath{\pm 0.079}}$ & $2.651 {\tinymath{\pm 0.045}}$ & $27.130 {\tinymath{\pm 0.224}}$ & $107.075 {\tinymath{\pm 1.398}}$ \\
                                & \method{IFTPP}    & $61.715 {\tinymath{\pm 0.152}}$ & $2.776 {\tinymath{\pm 0.043}}$ & $27.582 {\tinymath{\pm 0.191}}$ & $106.711 {\tinymath{\pm 1.615}}$ \\
                                & \method{TCDDM}    & $60.501 {\tinymath{\pm 0.087}}$ & $2.387 {\tinymath{\pm 0.050}}$ & $27.303 {\tinymath{\pm 0.152}}$ & $106.048 {\tinymath{\pm 0.610}}$ \\
                                & \method{CDiff}    & $60.661 {\tinymath{\pm 0.101}}$ & $\mathbf{2.293} {\tinymath{\pm 0.034}}$ & $27.101 {\tinymath{\pm 0.113}}$ & $106.184 {\tinymath{\pm 1.121}}$ \\
                                & \FIMzeroshot & $60.238$ \tinymath{\pm 0.161} & $4.172$ \tinymath{\pm 0.064} & $24.057$ \tinymath{\pm 0.050} & $99.069$ \tinymath{\pm 0.390}\\

                                & \FIMfine & $59.437$ \tinymath{\pm 0.082} & $2.703$ \tinymath{\pm 0.012} & $21.985$ \tinymath{\pm 0.014} & $87.585$ \tinymath{\pm 0.171}\\

                                & \textbf{\ourssynth} & $58.449$ \tinymath{\pm 0.000} & $7.277$ \tinymath{\pm 0.000} & $24.424$ \tinymath{\pm 0.000} & $98.867$ \tinymath{\pm 0.000} \\

                                & \textbf{\ours} & $\mathbf{57.966}$ \tinymath{\pm 0.000} & $6.102$ \tinymath{\pm 0.000} & $\mathbf{21.718}$ \tinymath{\pm 0.000} & $\mathbf{85.880}$ \tinymath{\pm 0.000} \\

\end{tabular}
\end{center}
\end{table}

\begin{table}[h]
\caption{Prediction of $N{=}10$ events on five real-world datasets. Baselines from \cite{cdiff,fim_pp}. Best in bold.}
\label{tab:long-horizon-pred-10}
\small
\begin{center}


\begin{tabular}{llrrrr}
Dataset              & Method   & \OTD                                   & \RMSEe                           & \RMSEdt                          & \SMAPEdt                                 \\
\midrule
\multirow{7}{*}{\datasetbf{Taxi}} & \method{HYPRO}    & $11.875 {\tinymath{\pm 0.172}}      $ & $0.764 {\tinymath{\pm 0.008}}$ & $0.363 {\tinymath{\pm 0.002}}$ & $89.524 {\tinymath{\pm 0.552}}$ \\
                       & \method{Dual-TPP} & $13.058 {\tinymath{\pm 0.220}}$ & $0.966 {\tinymath{\pm 0.011}}$ & $0.395 {\tinymath{\pm 0.003}}$ & $90.812 {\tinymath{\pm 0.497}}$ \\
                       & \method{A-NHP}    & $12.542 {\tinymath{\pm 0.336}}$ & $0.823 {\tinymath{\pm 0.007}}$ & $0.376 {\tinymath{\pm 0.003}}$ & $92.812 {\tinymath{\pm 0.129}}$ \\
                       & \method{NHP}      & $13.377 {\tinymath{\pm 0.184}}$ & $0.922 {\tinymath{\pm 0.009}}$ & $0.397 {\tinymath{\pm 0.005}}$ & $92.182 {\tinymath{\pm 0.384}}$ \\
                       & \method{IFTPP}    & $12.765 {\tinymath{\pm 0.106}}$ & $1.004 {\tinymath{\pm 0.013}}$ & $0.383 {\tinymath{\pm 0.015}}$ & $93.120 {\tinymath{\pm 0.526}}$ \\
                       & \method{TCDDM}    & $11.885 {\tinymath{\pm 0.149}}$ & $1.121 {\tinymath{\pm 0.072}}$ & $0.385 {\tinymath{\pm 0.009}}$ & $90.703 {\tinymath{\pm 0.356}}$ \\
                       & \method{CDiff}    & $11.004 {\tinymath{\pm 0.191}}$ & $0.785 {\tinymath{\pm 0.007}}$ & $0.350 {\tinymath{\pm 0.002}}$ & $90.721 {\tinymath{\pm 0.291}}$ \\
                       
                       & \FIMzeroshot& $13.820$ \tinymath{\pm 0.124} & $1.190$ \tinymath{\pm 0.013} & $0.281$ \tinymath{\pm 0.001} & $78.141$ \tinymath{\pm 0.414} \\
                       
                        & \FIMfine & $8.336$ \tinymath{\pm 0.071} & $\mathbf{0.451}$ \tinymath{\pm 0.006} & $0.291$ \tinymath{\pm 0.004} & $75.366$ \tinymath{\pm 0.160} \\

                        & \textbf{\ourssynth} & $9.753$ \tinymath{\pm 0.000} & $0.585$ \tinymath{\pm 0.000} & $0.273$ \tinymath{\pm 0.000} & $\mathbf{72.446}$ \tinymath{\pm 0.000} \\

                        & \textbf{\ours} & $\mathbf{8.324}$ \tinymath{\pm 0.000} & $0.453$ \tinymath{\pm 0.000} & $\mathbf{0.266}$ \tinymath{\pm 0.000} & $72.918$ \tinymath{\pm 0.000} \\

\midrule
\multirow{7}{*}{\datasetbf{Taobao}} & \method{HYPRO}    & $21.547 {\tinymath{\pm 0.138}}$ & $1.527 {\tinymath{\pm 0.035}}$ & $0.591 {\tinymath{\pm 0.019}}$ & $133.147 {\tinymath{\pm 0.341}}$ \\
                         & \method{Dual-TPP} & $23.691 {\tinymath{\pm 0.203}}$ & $2.674 {\tinymath{\pm 0.032}}$ & $0.873 {\tinymath{\pm 0.010}}$ & $139.271 {\tinymath{\pm 0.348}}$ \\
                         & \method{A-NHP}    & $21.683 {\tinymath{\pm 0.215}}$ & $1.514 {\tinymath{\pm 0.015}}$ & $0.608 {\tinymath{\pm 0.011}}$ & $135.271 {\tinymath{\pm 0.395}}$ \\
                         & \method{NHP}      & $24.068 {\tinymath{\pm 0.331}}$ & $2.769 {\tinymath{\pm 0.033}}$ & $0.855 {\tinymath{\pm 0.013}}$ & $137.693 {\tinymath{\pm 0.225}}$ \\
                         & \method{IFTPP}    & $23.195 {\tinymath{\pm 0.039}}$ & $2.429 {\tinymath{\pm 0.045}}$ & $0.602 {\tinymath{\pm 0.037}}$ & $127.411 {\tinymath{\pm 0.573}}$ \\
                         & \method{TCDDM}    & $\mathbf{21.012} {\tinymath{\pm 0.520}}$ & $2.598 {\tinymath{\pm 0.047}}$ & $0.610 {\tinymath{\pm 0.022}}$ & $132.711 {\tinymath{\pm 0.774}}$ \\
                         & \method{CDiff}    & $21.221 {\tinymath{\pm 0.176}}$ & $1.416 {\tinymath{\pm 0.024}}$ & $\mathbf{0.535} {\tinymath{\pm 0.016}}$ & $126.824 {\tinymath{\pm 0.366}}$ \\
                         
                         & \FIMzeroshot & $31.880$ \tinymath{\pm 0.040} & $2.024$ \tinymath{\pm 0.004} & $1.955$ \tinymath{\pm 0.011} & $170.278$ \tinymath{\pm 0.029}\\
                          
                         & \FIMfine & $27.974$ \tinymath{\pm 0.162} & $\mathbf{1.325}$ \tinymath{\pm 0.010} & $14.954$ \tinymath{\pm 0.253} & $145.821$ \tinymath{\pm 1.120}\\

                         & \textbf{\ourssynth} & $22.056$ \tinymath{\pm 0.000} & $1.498$ \tinymath{\pm 0.000} & $1.244$ \tinymath{\pm 0.000} & $\mathbf{121.642}$ \tinymath{\pm 0.000} \\

                         & \textbf{\ours} & $23.928$ \tinymath{\pm 0.000} & $1.518$ \tinymath{\pm 0.000} & $1.414$ \tinymath{\pm 0.000} & $168.930$ \tinymath{\pm 0.000} \\
    
\midrule
\multirow{7}{*}{\datasetbf{StackOverflow}} & \method{HYPRO}    & $21.062 {\tinymath{\pm 0.372}}      $ & $0.921 {\tinymath{\pm 0.019}}      $ & $1.235 {\tinymath{\pm 0.006}}$ & $107.566 {\tinymath{\pm 0.218}}$ \\
                                & \method{Dual-TPP} & $21.229 {\tinymath{\pm 0.394}}$ & $0.936 {\tinymath{\pm 0.013}}$ & $1.223 {\tinymath{\pm 0.010}}$ & $107.274 {\tinymath{\pm 0.200}}$ \\
                                & \method{A-NHP}    & $22.019 {\tinymath{\pm 0.220}}$ & $0.978 {\tinymath{\pm 0.023}}$ & $1.225 {\tinymath{\pm 0.007}}$ & $100.137 {\tinymath{\pm 0.167}}$ \\
                                & \method{NHP}      & $21.655 {\tinymath{\pm 0.314}}$ & $0.970 {\tinymath{\pm 0.014}}$ & $1.266 {\tinymath{\pm 0.003}}$ & $108.867 {\tinymath{\pm 0.361}}$ \\
                                & \method{IFTPP}    & $22.339 {\tinymath{\pm 0.322}}$ & $0.970 {\tinymath{\pm 0.011}}$ & $1.251 {\tinymath{\pm 0.005}}$ & $105.674 {\tinymath{\pm 0.337}}$ \\
                                & \method{TCDDM}    & $22.042 {\tinymath{\pm 0.193}}$ & $1.205 {\tinymath{\pm 0.014}}$ & $1.228 {\tinymath{\pm 0.010}}$ & $108.111 {\tinymath{\pm 0.112}}$ \\
                                & \method{CDiff}    & $20.191 {\tinymath{\pm 0.455}}$ & $0.916 {\tinymath{\pm 0.010}}$ & $1.180 {\tinymath{\pm 0.003}}$ & $102.367 {\tinymath{\pm 0.267}}$ \\
                                
                                & \FIMzeroshot & $23.527$ \tinymath{\pm 0.033} & $1.188$ \tinymath{\pm 0.005} & $1.039$ \tinymath{\pm 0.003} & $92.919$ \tinymath{\pm 0.556} \\   
                                
                                &\FIMfine & $\mathbf{19.938}$ \tinymath{\pm 0.093} & $\mathbf{0.823}$ \tinymath{\pm 0.010} & $1.012$ \tinymath{\pm 0.004} & $87.503$ \tinymath{\pm 0.402} \\

                                & \textbf{\ourssynth} & $24.075$ \tinymath{\pm 0.000} & $1.463$ \tinymath{\pm 0.000} & $0.991$ \tinymath{\pm 0.000} & $87.078$ \tinymath{\pm 0.000} \\

                                & \textbf{\ours} & $23.075$ \tinymath{\pm 0.000} & $1.445$ \tinymath{\pm 0.000} & $\mathbf{0.960}$ \tinymath{\pm 0.000} & $\mathbf{85.819}$ \tinymath{\pm 0.000} \\

\midrule
\multirow{7}{*}{\datasetbf{Amazon}}        & \method{HYPRO}    & $24.956 {\tinymath{\pm 0.663}}$ & $1.765 {\tinymath{\pm 0.039}}      $ & $0.442 {\tinymath{\pm 0.015}}$ & $83.401 {\tinymath{\pm 1.033}}      $ \\
                                & \method{Dual-TPP} & $25.929 {\tinymath{\pm 0.280}}$ & $2.098 {\tinymath{\pm 0.101}}$ & $0.475 {\tinymath{\pm 0.008}}$ & $82.352 {\tinymath{\pm 1.285}}$ \\
                                & \method{A-NHP}    & $24.116 {\tinymath{\pm 0.807}}$ & $1.741 {\tinymath{\pm 0.039}}$ & $0.454 {\tinymath{\pm 0.014}}$ & $84.323 {\tinymath{\pm 1.815}}$ \\
                                & \method{NHP}      & $25.730 {\tinymath{\pm 0.497}}$ & $1.843 {\tinymath{\pm 0.053}}$ & $0.491 {\tinymath{\pm 0.048}}$ & $89.135 {\tinymath{\pm 1.092}}$ \\
                                & \method{IFTPP}    & $26.632 {\tinymath{\pm 0.519}}$ & $1.955 {\tinymath{\pm 0.112}}$ & $0.464 {\tinymath{\pm 0.066}}$ & $89.305 {\tinymath{\pm 1.288}}$ \\
                                & \method{TCDDM}    & $25.091 {\tinymath{\pm 0.227}}$ & $1.778 {\tinymath{\pm 0.090}}$ & $0.448 {\tinymath{\pm 0.082}}$ & $82.105 {\tinymath{\pm 1.564}}$ \\
                                & \method{CDiff}    & $24.230 {\tinymath{\pm 0.287}}$ & $1.766 {\tinymath{\pm 0.079}}$ & $0.450 {\tinymath{\pm 0.049}}$ & $82.124 {\tinymath{\pm 2.094}}$ \\
                                
                                & \FIMzeroshot & $21.736$ \tinymath{\pm 0.115} & $1.141$ \tinymath{\pm 0.010} & $0.449$ \tinymath{\pm 0.002} & $120.894$ \tinymath{\pm 0.393}\\
                                
                                & \FIMfine& $\mathbf{18.428}$ \tinymath{\pm 0.124} & $\mathbf{1.091}$ \tinymath{\pm 0.016} & $0.361$ \tinymath{\pm 0.001} & $87.264$ \tinymath{\pm 0.323}\\

                                & \textbf{\ourssynth} & $22.840$ \tinymath{\pm 0.000} & $1.885$ \tinymath{\pm 0.000} & $0.324$ \tinymath{\pm 0.000} & $\mathbf{59.647}$ \tinymath{\pm 0.000} \\

                                & \textbf{\ours} & $21.947$ \tinymath{\pm 0.000} & $1.701$ \tinymath{\pm 0.000} & $0.314$ \tinymath{\pm 0.000} & $74.389$ \tinymath{\pm 0.000} \\
\midrule
\multirow{7}{*}{\datasetbf{Retweet}}       & \method{HYPRO}    & $31.743 {\tinymath{\pm 0.068}}$ & $1.927 {\tinymath{\pm 0.027}}$ & $33.683 {\tinymath{\pm 0.245}}$ & $105.073 {\tinymath{\pm 0.958}}      $ \\
                                & \method{Dual-TPP} & $31.652 {\tinymath{\pm 0.075}}$ & $1.963 {\tinymath{\pm 0.038}}$ & $28.104 {\tinymath{\pm 0.486}}$ & $106.721 {\tinymath{\pm 0.774}}$ \\
                                & \method{A-NHP}    & $30.337 {\tinymath{\pm 0.065}}$ & $1.823 {\tinymath{\pm 0.031}}$ & $26.310 {\tinymath{\pm 0.333}}$ & $106.021 {\tinymath{\pm 1.011}}$ \\
                                & \method{NHP}      & $30.817 {\tinymath{\pm 0.090}}$ & $1.713 {\tinymath{\pm 0.024}}$ & $27.010 {\tinymath{\pm 0.429}}$ & $107.053 {\tinymath{\pm 1.390}}$ \\
                                & \method{IFTPP}    & $31.974 {\tinymath{\pm 0.032}}$ & $1.942 {\tinymath{\pm 0.062}}$ & $28.825 {\tinymath{\pm 0.221}}$ & $106.014 {\tinymath{\pm 0.633}}$ \\
                                & \method{TCDDM}    & $32.006 {\tinymath{\pm 0.074}}$ & $1.789 {\tinymath{\pm 0.094}}$ & $29.124 {\tinymath{\pm 0.405}}$ & $106.738 {\tinymath{\pm 0.791}}$ \\
                                & \method{CDiff}    & $31.237 {\tinymath{\pm 0.078}}$ & $1.745 {\tinymath{\pm 0.036}}$ & $26.429 {\tinymath{\pm 0.201}}$ & $105.767 {\tinymath{\pm 0.771}}$ \\
                                
                                & \FIMzeroshot & $31.027$ \tinymath{\pm 0.031} & $2.355$ \tinymath{\pm 0.032} & $27.085$ \tinymath{\pm 0.002} & $97.590$ \tinymath{\pm 0.152}\\                                                               
                                & \FIMfine& $30.592$ \tinymath{\pm 0.037} & $\mathbf{1.611}$ \tinymath{\pm 0.031} & $25.021$ \tinymath{\pm 0.034} & $86.875$ \tinymath{\pm 0.108}\\

                                & \textbf{\ourssynth} & $30.539$ \tinymath{\pm 0.000} & $3.769$ \tinymath{\pm 0.000} & $27.525$ \tinymath{\pm 0.000} & $98.648$ \tinymath{\pm 0.000} \\

                                & \textbf{\ours} & $\mathbf{30.330}$ \tinymath{\pm 0.000} & $3.116$ \tinymath{\pm 0.000} & $\mathbf{24.819}$ \tinymath{\pm 0.000} & $\mathbf{84.946}$ \tinymath{\pm 0.000} \\
\end{tabular}
\end{center}
\end{table}

\begin{table}[h]
\caption{Prediction of $N{=}5$ events on five real-world datasets (full results). Baselines from \cite{cdiff,fim_pp}. Best in bold.}
\label{tab:long-horizon-pred-5}
\small
\begin{center}


\begin{tabular}{llrrrr}
Dataset              & Method   & \OTD                                   & \RMSEe                           & \RMSEdt                          & \SMAPEdt                                 \\
\midrule
\multirow{7}{*}{\datasetbf{Taxi}} & \method{HYPRO}    & $5.952 {\tinymath{\pm 0.126}}$ & $0.500 {\tinymath{\pm 0.011}}$ & $0.322 {\tinymath{\pm 0.004}}$ & $85.994 {\tinymath{\pm 0.227}}$ \\
                       & \method{Dual-TPP} & $7.534 {\tinymath{\pm 0.111}}$ & $0.636 {\tinymath{\pm 0.009}}$ & $0.340 {\tinymath{\pm 0.003}}$ & $89.727 {\tinymath{\pm 0.320}}$ \\
                       & \method{A-NHP}    & $6.441 {\tinymath{\pm 0.090}}$ & $0.682 {\tinymath{\pm 0.010}}$ & $0.347 {\tinymath{\pm 0.002}}$ & $89.070 {\tinymath{\pm 0.152}}$ \\
                       & \method{NHP}      & $7.405 {\tinymath{\pm 0.122}}$ & $0.641 {\tinymath{\pm 0.013}}$ & $0.351 {\tinymath{\pm 0.008}}$ & $91.625 {\tinymath{\pm 0.177}}$ \\
                       & \method{IFTPP}    & $7.209 {\tinymath{\pm 0.184}}$ & $0.608 {\tinymath{\pm 0.008}}$ & $0.335 {\tinymath{\pm 0.003}}$ & $90.512 {\tinymath{\pm 0.169}}$ \\
                       & \method{TCDDM}    & $5.877 {\tinymath{\pm 0.095}}$ & $0.648 {\tinymath{\pm 0.015}}$ & $0.327 {\tinymath{\pm 0.005}}$ & $88.051 {\tinymath{\pm 0.240}}$ \\
                       & \method{CDiff}    & $5.966 {\tinymath{\pm 0.083}}$ & $0.547 {\tinymath{\pm 0.007}}$ & $0.318 {\tinymath{\pm 0.003}}$ & $89.535 {\tinymath{\pm 0.294}}$ \\
                       & \FIMzeroshot & $6.773$ \tinymath{\pm 0.064} & $0.655$ \tinymath{\pm 0.013} & $0.246$ \tinymath{\pm 0.001} & $74.912$ \tinymath{\pm 0.793} \\
                    
                       & \FIMfine & $4.083$ \tinymath{\pm 0.032} & $0.311$ \tinymath{\pm 0.007} & $0.250$ \tinymath{\pm 0.002} & $71.108$ \tinymath{\pm 0.902} \\

                       & \textbf{\ourssynth} & $4.385$ \tinymath{\pm 0.000} & $0.343$ \tinymath{\pm 0.000} & $\mathbf{0.236}$ \tinymath{\pm 0.000} & $\mathbf{69.672}$ \tinymath{\pm 0.000} \\

                       & \textbf{\ours} & $\mathbf{4.054}$ \tinymath{\pm 0.000} & $\mathbf{0.299}$ \tinymath{\pm 0.000} & $0.241$ \tinymath{\pm 0.000} & $71.038$ \tinymath{\pm 0.000} \\

\midrule
\multirow{7}{*}{\datasetbf{Taobao}} & \method{HYPRO}    & $11.317 {\tinymath{\pm 0.111}}$ & $0.817 {\tinymath{\pm 0.037}}$ & $0.573 {\tinymath{\pm 0.011}}$ & $133.837 {\tinymath{\pm 0.524}}$ \\
                         & \method{Dual-TPP} & $13.280 {\tinymath{\pm 0.092}}$ & $1.877 {\tinymath{\pm 0.014}}$ & $0.691 {\tinymath{\pm 0.007}}$ & $134.437 {\tinymath{\pm 0.458}}$ \\
                         & \method{A-NHP}    & $11.223 {\tinymath{\pm 0.145}}$ & $0.873 {\tinymath{\pm 0.023}}$ & $0.550 {\tinymath{\pm 0.014}}$ & $132.266 {\tinymath{\pm 0.532}}$ \\
                         & \method{NHP}      & $11.973 {\tinymath{\pm 0.176}}$ & $1.910 {\tinymath{\pm 0.031}}$ & $0.712 {\tinymath{\pm 0.017}}$ & $134.693 {\tinymath{\pm 0.369}}$ \\
                         & \method{IFTPP}    & $11.052 {\tinymath{\pm 0.108}}$ & $1.941 {\tinymath{\pm 0.049}}$ & $0.601 {\tinymath{\pm 0.017}}$ & $126.320 {\tinymath{\pm 0.591}}$ \\
                         & \method{TCDDM}    & $11.609 {\tinymath{\pm 0.184}}$ & $1.690 {\tinymath{\pm 0.023}}$ & $0.675 {\tinymath{\pm 0.009}}$ & $129.009 {\tinymath{\pm 0.923}}$ \\
                         & \method{CDiff}    & $10.147 {\tinymath{\pm 0.140}}$ & $\mathbf{0.730} {\tinymath{\pm 0.019}}$ & $\mathbf{0.519} {\tinymath{\pm 0.008}}$ & $124.339 {\tinymath{\pm 0.322}}$ \\
                         & \FIMzeroshot & $15.951$ \tinymath{\pm 0.042} & $1.129$ \tinymath{\pm 0.007} & $1.761$ \tinymath{\pm 0.013} & $168.299$ \tinymath{\pm 0.249} \\
                         
                         & \FIMfine & $13.173$ \tinymath{\pm 0.261} & $0.745$ \tinymath{\pm 0.010} & $14.892$ \tinymath{\pm 0.370} & $146.921$ \tinymath{\pm 0.858}\\

                         & \textbf{\ourssynth} & $\mathbf{9.894}$ \tinymath{\pm 0.000} & $\mathbf{0.730}$ \tinymath{\pm 0.000} & $0.974$ \tinymath{\pm 0.000} & $\mathbf{121.309}$ \tinymath{\pm 0.000} \\

                         & \textbf{\ours} & $10.938$ \tinymath{\pm 0.000} & $0.745$ \tinymath{\pm 0.000} & $1.167$ \tinymath{\pm 0.000} & $166.051$ \tinymath{\pm 0.000} \\

\midrule
\multirow{7}{*}{\datasetbf{StackOverflow}} & \method{HYPRO}    & $11.590 {\tinymath{\pm 0.186}}$ & $0.586 {\tinymath{\pm 0.019}}$ & $1.227 {\tinymath{\pm 0.018}}$ & $109.014 {\tinymath{\pm 0.422}}$ \\
                                & \method{Dual-TPP} & $11.719 {\tinymath{\pm 0.109}}$ & $0.591 {\tinymath{\pm 0.026}}$ & $1.296 {\tinymath{\pm 0.010}}$ & $106.697 {\tinymath{\pm 0.381}}$ \\
                                & \method{A-NHP}    & $11.595 {\tinymath{\pm 0.197}}$ & $0.575 {\tinymath{\pm 0.009}}$ & $1.188 {\tinymath{\pm 0.014}}$ & $105.799 {\tinymath{\pm 0.516}}$ \\
                                & \method{NHP}      & $11.807 {\tinymath{\pm 0.155}}$ & $0.596 {\tinymath{\pm 0.015}}$ & $1.261 {\tinymath{\pm 0.013}}$ & $108.074 {\tinymath{\pm 0.661}}$ \\
                                & \method{IFTPP}    & $13.124 {\tinymath{\pm 0.174}}$ & $0.702 {\tinymath{\pm 0.008}}$ & $1.182 {\tinymath{\pm 0.039}}$ & $108.409 {\tinymath{\pm 0.692}}$ \\
                                & \method{TCDDM}    & $11.410 {\tinymath{\pm 0.129}}$ & $0.630 {\tinymath{\pm 0.015}}$ & $1.201 {\tinymath{\pm 0.028}}$ & $107.893 {\tinymath{\pm 0.942}}$ \\
                                & \method{CDiff}    & $10.735 {\tinymath{\pm 0.183}}$ & $0.571 {\tinymath{\pm 0.012}}$ & $1.153 {\tinymath{\pm 0.011}}$ & $100.586 {\tinymath{\pm 0.299}}$ \\
                                & \FIMzeroshot & $11.520$ \tinymath{\pm 0.057} & $0.657$ \tinymath{\pm 0.003} & $1.030$ \tinymath{\pm 0.001} & $93.296$ \tinymath{\pm 0.506} \\
                               
                                & \FIMfine & $\mathbf{10.353}$ \tinymath{\pm 0.051} & $\mathbf{0.527}$ \tinymath{\pm 0.004} & $0.990$ \tinymath{\pm 0.003} & $86.443$ \tinymath{\pm 0.128} \\

                                & \textbf{\ourssynth} & $11.913$ \tinymath{\pm 0.000} & $0.773$ \tinymath{\pm 0.000} & $0.972$ \tinymath{\pm 0.000} & $86.366$ \tinymath{\pm 0.000} \\

                                & \textbf{\ours} & $11.645$ \tinymath{\pm 0.000} & $0.772$ \tinymath{\pm 0.000} & $\mathbf{0.946}$ \tinymath{\pm 0.000} & $\mathbf{84.368}$ \tinymath{\pm 0.000} \\

\midrule
\multirow{7}{*}{\datasetbf{Amazon}} & \method{HYPRO}    & $9.552 {\tinymath{\pm 0.172}}$ & $1.397 {\tinymath{\pm 0.033}}$ & $0.433 {\tinymath{\pm 0.008}}$ & $82.847 {\tinymath{\pm 0.748}}$ \\
                         & \method{Dual-TPP} & $11.309 {\tinymath{\pm 0.093}}$ & $1.742 {\tinymath{\pm 0.302}}$ & $0.476 {\tinymath{\pm 0.010}}$ & $86.633 {\tinymath{\pm 0.573}}$ \\
                         & \method{A-NHP}    & $\mathbf{9.430} {\tinymath{\pm 0.131}}$ & $1.117 {\tinymath{\pm 0.049}}$ & $0.427 {\tinymath{\pm 0.033}}$ & $83.121 {\tinymath{\pm 0.415}}$ \\
                         & \method{NHP}      & $11.273 {\tinymath{\pm 0.198}}$ & $1.431 {\tinymath{\pm 0.024}}$ & $0.501 {\tinymath{\pm 0.009}}$ & $90.591 {\tinymath{\pm 0.667}}$ \\
                         & \method{IFTPP}    & $10.230 {\tinymath{\pm 0.224}}$ & $1.663 {\tinymath{\pm 0.168}}$ & $0.447 {\tinymath{\pm 0.015}}$ & $88.900 {\tinymath{\pm 0.610}}$ \\
                         & \method{TCDDM}    & $10.557 {\tinymath{\pm 0.331}}$ & $1.409 {\tinymath{\pm 0.203}}$ & $0.460 {\tinymath{\pm 0.032}}$ & $82.401 {\tinymath{\pm 0.810}}$ \\
                         & \method{CDiff}    & $9.478 {\tinymath{\pm 0.081}}$ & $1.326 {\tinymath{\pm 0.082}}$ & $0.424 {\tinymath{\pm 0.018}}$ & $81.287 {\tinymath{\pm 0.994}}$ \\
                         & \FIMzeroshot & $11.124$ \tinymath{\pm 0.059} & $\mathbf{0.736}$ \tinymath{\pm 0.004} & $0.449$ \tinymath{\pm 0.004} & $119.129$ \tinymath{\pm 0.746} \\
                         
                         & \FIMfine & $10.034$ \tinymath{\pm 0.060} & $0.737$ \tinymath{\pm 0.006} & $0.341$ \tinymath{\pm 0.004} & $78.738$ \tinymath{\pm 0.339}\\

                         & \textbf{\ourssynth} & $11.459$ \tinymath{\pm 0.000} & $0.982$ \tinymath{\pm 0.000} & $\mathbf{0.289}$ \tinymath{\pm 0.000} & $\mathbf{57.904}$ \tinymath{\pm 0.000} \\

                         & \textbf{\ours} & $10.930$ \tinymath{\pm 0.000} & $0.894$ \tinymath{\pm 0.000} & $0.297$ \tinymath{\pm 0.000} & $71.182$ \tinymath{\pm 0.000} \\

\midrule
\multirow{7}{*}{\datasetbf{Retweet}} & \method{HYPRO}    & $16.145 {\tinymath{\pm 0.096}}$ & $1.105 {\tinymath{\pm 0.026}}$ & $27.236 {\tinymath{\pm 0.259}}$ & $103.052 {\tinymath{\pm 1.206}}$ \\
                         & \method{Dual-TPP} & $16.050 {\tinymath{\pm 0.085}}$ & $1.077 {\tinymath{\pm 0.027}}$ & $31.493 {\tinymath{\pm 0.162}}$ & $101.322 {\tinymath{\pm 1.127}}$ \\
                         & \method{A-NHP}    & $16.124 {\tinymath{\pm 0.089}}$ & $1.058 {\tinymath{\pm 0.029}}$ & $29.247 {\tinymath{\pm 0.145}}$ & $105.930 {\tinymath{\pm 1.380}}$ \\
                         & \method{NHP}      & $15.945 {\tinymath{\pm 0.094}}$ & $1.113 {\tinymath{\pm 0.040}}$ & $32.367 {\tinymath{\pm 0.104}}$ & $107.022 {\tinymath{\pm 1.077}}$ \\
                         & \method{IFTPP}    & $16.043 {\tinymath{\pm 0.222}}$ & $1.313 {\tinymath{\pm 0.011}}$ & $30.853 {\tinymath{\pm 0.119}}$ & $106.941 {\tinymath{\pm 2.031}}$ \\
                         & \method{TCDDM}    & $15.874 {\tinymath{\pm 0.053}}$ & $1.194 {\tinymath{\pm 0.021}}$ & $28.530 {\tinymath{\pm 0.110}}$ & $105.570 {\tinymath{\pm 0.940}}$ \\
                         & \method{CDiff}    & $15.858 {\tinymath{\pm 0.080}}$ & $\mathbf{1.023} {\tinymath{\pm 0.036}}$ & $26.078 {\tinymath{\pm 0.175}}$ & $106.620 {\tinymath{\pm 1.008}}$ \\
                         & \FIMzeroshot & $15.747$ \tinymath{\pm 0.032} & $1.342$ \tinymath{\pm 0.027} & $28.138$ \tinymath{\pm 0.068} & $98.668$ \tinymath{\pm 0.794}\\
                         
                         & \FIMfine & $15.645$ \tinymath{\pm 0.020} & $1.033$ \tinymath{\pm 0.034} & $25.308$ \tinymath{\pm 0.135} & $83.010$ \tinymath{\pm 0.278}\\

                         & \textbf{\ourssynth} & $15.577$ \tinymath{\pm 0.000} & $1.908$ \tinymath{\pm 0.000} & $27.093$ \tinymath{\pm 0.000} & $92.250$ \tinymath{\pm 0.000} \\

                         & \textbf{\ours} & $\mathbf{15.557}$ \tinymath{\pm 0.000} & $1.603$ \tinymath{\pm 0.000} & $\mathbf{24.738}$ \tinymath{\pm 0.000} & $\mathbf{82.464}$ \tinymath{\pm 0.000} \\
           
\end{tabular}

\end{center}
\end{table}

\FloatBarrier

\section{Ablations}
\label{app:ablations}
\subsection{Dataset-specific training}
As an ablation study, we also evaluated what happens when evolution is performed on only a single dataset. The goal is to test whether the model can discover an algorithm that performs better on that specific dataset and to what extent such an algorithm still generalizes to the others. Table~\ref{tab:tpp-one-dataset-evolution} shows the results for evolving only on Amazon. We find that on Amazon this dataset-specific variant improves slightly in $\text{sMAPE}_{\Delta t}$, while its \OTD is slightly worse. At the same time, the evolved algorithm still generalizes reasonably well to the other datasets. Overall, this suggests that, at least in our setting where the model does not have direct access to the data through plots or other auxiliary views, evolving on only a single dataset does not lead to significant differences.
\begin{table}[t]
    \caption{Long-horizon prediction ($N{=}5$) on five real-world datasets. \ours(Amazon-only) has only been evolved on the amazon dataset. We show OTD and $\text{sMAPE}_{\Delta t}$ (both lower is better) for the best neural baseline (CDiff), FIM-PP variants, and \ours variants. Baseline results from \cite{cdiff,fim_pp}. FIM(zs) is FIM-PP \cite{fim_pp} in zero-shot mode, FIM(f) is FIM-PP after finetuning on the target dataset. EVIL(s) is \ourssynth.}
    \label{tab:tpp-one-dataset-evolution}
    \scriptsize
    \centering
    \setlength{\tabcolsep}{3pt}
    \begin{tabular}{l cc cc cc cc cc}
    \toprule
     & \multicolumn{2}{c}{\datasetbf{Taxi}} & \multicolumn{2}{c}{\datasetbf{Taobao}} & \multicolumn{2}{c}{\datasetbf{StackOv.}} & \multicolumn{2}{c}{\datasetbf{Amazon}} & \multicolumn{2}{c}{\datasetbf{Retweet}} \\
    \cmidrule(lr){2-3}\cmidrule(lr){4-5}\cmidrule(lr){6-7}\cmidrule(lr){8-9}\cmidrule(lr){10-11}
    Method & OTD & sM & OTD & sM & OTD & sM & OTD & sM & OTD & sM \\
    \midrule
    CDiff & $5.97$ & $89.5$ & $10.15$ & $124.3$ & $10.74$ & $100.6$ & $\mathbf{9.48}$ & $81.3$ & $15.86$ & $106.6$ \\
    FIM(zs) & $6.77$ & $74.9$ & $15.95$ & $168.3$ & $11.52$ & $93.3$ & $11.12$ & $119.1$ & $15.75$ & $98.7$ \\
    FIM(f) & $4.08$ & $71.1$ & $13.17$ & $146.9$ & $\mathbf{10.35}$ & $86.4$ & $10.03$ & $78.7$ & $15.65$ & $83.0$ \\
    \midrule
    \textbf{EVIL(s)} & $4.39$ & $\mathbf{69.7}$ & $\mathbf{9.89}$ & $\mathbf{121.3}$ & $11.91$ & $86.4$ & $11.46$ & $\mathbf{57.9}$ & $15.58$ & $92.3$ \\
    \textbf{EVIL} & $\mathbf{4.05}$ & $71.0$ & $10.94$ & $166.1$ & $11.65$ & $\mathbf{84.4}$ & $10.93$ & $71.2$ & $15.56$ & $\mathbf{82.5}$ \\
    \textbf{EVIL(amazon only)} & $\mathbf{3.93}$ & $72.5$ & $10.24$ & $\mathbf{119.8}$ & $12.07$ & $86.3$ & $11.16$ & $\mathbf{54.7}$ & $15.57$ & $98.7$ \\
    \bottomrule
    \end{tabular}
    \vspace{-8pt}
    \end{table}

\subsection{Variance between independent runs}
To assess how sensitive the evolutionary search is to the random seed, we repeated the same setup across five independent runs with different seeds. Table~\ref{tab:tpp-independent-runs} reports the mean and standard deviation across these runs. We observe moderate variability across datasets and metrics, which is expected for an evolutionary search procedure. At the same time, the mean results remain broadly competitive with the baselines, and the standard deviations are generally small relative to the gaps between methods. This suggests that the method is not entirely seed-invariant, but its overall performance profile is fairly stable.

\begin{table}[t]
    \caption{Long-horizon prediction ($N{=}5$) on five real-world datasets: mean $\pm$ std over five independent \ours runs with different seeds. We show OTD and $\text{sMAPE}_{\Delta t}$ (both lower is better) together with the strongest neural baseline (CDiff) and FIM-PP variants. Baseline results from \cite{cdiff,fim_pp}. FIM(zs) is FIM-PP \cite{fim_pp} in zero-shot mode, and FIM(f) is FIM-PP after finetuning on the target dataset.}
    \label{tab:tpp-independent-runs}
    \scriptsize
    \begin{center}
    \setlength{\tabcolsep}{3pt}
    \resizebox{\linewidth}{!}{%
    \begin{tabular}{l cc cc cc cc cc}
    \toprule
     & \multicolumn{2}{c}{\datasetbf{Taxi}} & \multicolumn{2}{c}{\datasetbf{Taobao}} & \multicolumn{2}{c}{\datasetbf{StackOv.}} & \multicolumn{2}{c}{\datasetbf{Amazon}} & \multicolumn{2}{c}{\datasetbf{Retweet}} \\
    \cmidrule(lr){2-3}\cmidrule(lr){4-5}\cmidrule(lr){6-7}\cmidrule(lr){8-9}\cmidrule(lr){10-11}
    Method & OTD & sM & OTD & sM & OTD & sM & OTD & sM & OTD & sM \\
    \midrule
    CDiff & $5.97$ & $89.5$ & $10.15$ & $124.3$ & $10.74$ & $100.6$ & $\mathbf{9.48}$ & $81.3$ & $15.86$ & $106.6$ \\
    \midrule
    FIM(zs) & $6.77$ & $74.9$ & $15.95$ & $168.3$ & $11.52$ & $93.3$ & $11.12$ & $119.1$ & $15.75$ & $98.7$ \\
    FIM(f) & $4.08$ & $71.1$ & $\mathbf{10.03}$ & $146.9$ & $\mathbf{10.35}$ & $86.4$ & $10.03$ & $78.7$ & $15.65$ & $83.0$ \\
    \midrule
    \textbf{EVIL} & $4.08 {\tinymath{\pm 0.13}}$ & $71.7 {\tinymath{\pm 3.3}}$ & $10.28 {\tinymath{\pm 0.33}}$ & $131.8 {\tinymath{\pm 17.7}}$ & $11.87 {\tinymath{\pm 0.15}}$ & $\mathbf{85.5} {\tinymath{\pm 1.3}}$ & $11.36 {\tinymath{\pm 0.25}}$ & $\mathbf{62.0} {\tinymath{\pm 6.3}}$ & $\mathbf{15.57} {\tinymath{\pm 0.07}}$ & $\mathbf{89.5} {\tinymath{\pm 6.2}}$ \\
    \bottomrule
    \end{tabular}%
    }
    \end{center}
    \vspace{-8pt}
    \end{table}

\subsection{Performance of a long run}
\begin{figure}[t]
    \centering
    \includegraphics[width=0.8\linewidth]{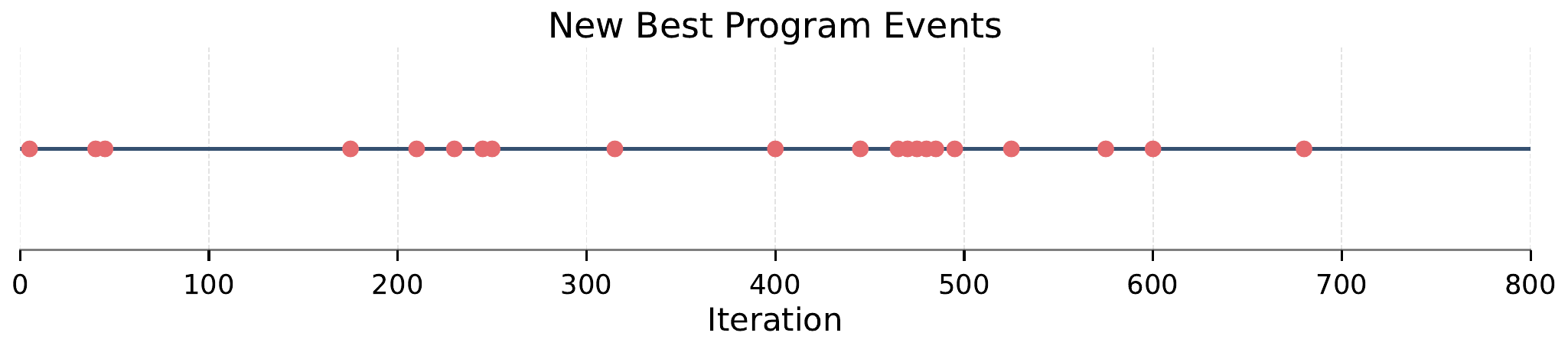}
    \caption{New best validation-set programs discovered during the 800-iteration long run. Across the run, the search found 21 new best algorithms, indicating continued progress on the validation objective even though the final gains on the benchmark test datasets remain small.}
    \label{fig:tpp-long-run-new-best}
    \vspace{-8pt}
\end{figure}

We also ran a substantially longer evolution with 800 iterations instead of 100 to test whether extended search would discover qualitatively new algorithms. The resulting test-set performance is summarized in Table~\ref{tab:tpp-long-run}. In this longer run, the search produced 21 new best algorithms on the validation set, as shown in Figure~\ref{fig:tpp-long-run-new-best}, and the final discovered program was much more involved, with roughly 400 lines of code. This indicates that the search continues to explore new parts of the program space over time. However, this added complexity only resulted in tiny improvements on the validation set and no noticeable improvements on the test set. This suggests that, at least in our current setup, the search is able to find most of the performance gains within the first 100 iterations, and further search does not lead to significant improvements on the benchmark datasets. We hypothesize that this could be due to a combination of factors, including the limited size of the validation set, the inherent noise in the evaluation process, and the fact that the benchmark datasets may not be sufficiently challenging to differentiate between closely related algorithms. Additionally, we set the time limit to 5 minutes per run which means that some potentially very advanced well performing algorithms might have been discarded.
\begin{table}[t]
    \caption{Long-horizon prediction ($N{=}5$) on five real-world datasets for a longer \ours run with 800 iterations. We show OTD and $\text{sMAPE}_{\Delta t}$ (both lower is better) together with the strongest neural baseline (CDiff), FIM-PP variants, and the standard \ours variants. Baseline results from \cite{cdiff,fim_pp}. FIM(zs) is FIM-PP \cite{fim_pp} in zero-shot mode, FIM(f) is FIM-PP after finetuning on the target dataset, and EVIL(s) is \ourssynth.}
    \label{tab:tpp-long-run}
    \scriptsize
    \centering
    \setlength{\tabcolsep}{3pt}
    \begin{tabular}{l cc cc cc cc cc}
    \toprule
     & \multicolumn{2}{c}{\datasetbf{Taxi}} & \multicolumn{2}{c}{\datasetbf{Taobao}} & \multicolumn{2}{c}{\datasetbf{StackOv.}} & \multicolumn{2}{c}{\datasetbf{Amazon}} & \multicolumn{2}{c}{\datasetbf{Retweet}} \\
    \cmidrule(lr){2-3}\cmidrule(lr){4-5}\cmidrule(lr){6-7}\cmidrule(lr){8-9}\cmidrule(lr){10-11}
    Method & OTD & sM & OTD & sM & OTD & sM & OTD & sM & OTD & sM \\
    \midrule
    CDiff & $5.97$ & $89.5$ & $10.15$ & $124.3$ & $10.74$ & $100.6$ & $\mathbf{9.48}$ & $81.3$ & $15.86$ & $106.6$ \\
    \midrule
    FIM(zs) & $6.77$ & $74.9$ & $15.95$ & $168.3$ & $11.52$ & $93.3$ & $11.12$ & $119.1$ & $15.75$ & $98.7$ \\
    FIM(f) & $4.08$ & $71.1$ & $13.17$ & $146.9$ & $\mathbf{10.35}$ & $86.4$ & $10.03$ & $78.7$ & $15.65$ & $83.0$ \\
    \midrule
    \textbf{EVIL(s)} & $4.39$ & $\mathbf{69.7}$ & $\mathbf{9.89}$ & $\mathbf{121.3}$ & $11.91$ & $86.4$ & $11.46$ & $\mathbf{57.9}$ & $15.58$ & $92.3$ \\
    \textbf{EVIL} & $\mathbf{4.05}$ & $71.0$ & $10.94$ & $166.1$ & $11.65$ & $\mathbf{84.4}$ & $10.93$ & $71.2$ & $15.56$ & $\mathbf{82.5}$ \\
    \textbf{EVIL (800 iterations)} & $4.06$ & $71.6$ & $9.93$ & $125.0$ & $11.90$ & $86.3$ & $11.36$ & $67.8$ & $\mathbf{15.40}$ & $90.8$ \\
    \bottomrule
    \end{tabular}
    \vspace{-8pt}
    \end{table}

\section{Additional Markov Jump Process Details}
\label{app:mjp_details}

\subsection{Continuous-Time Markov Jump Processes}

An MJP is a continuous-time Markov chain with right-continuous, piecewise-constant trajectories on a finite state space \cite{norris1998markov}. It is parameterized by an initial distribution $\pi_0$ and a generator (rate matrix) $Q \in \mathbb{R}^{K \times K}$ satisfying
\begin{equation}
Q_{ij} \geq 0 \ \text{for } i \neq j,
\qquad
Q_{ii} = - \sum_{j \neq i} Q_{ij}.
\end{equation}
The off-diagonal entry $Q_{ij}$ is the instantaneous jump rate from state $i$ to state $j$, while the diagonal term sets the total exit rate from state $i$. If we write the state-occupation probabilities as a row vector $p(t)$, then the dynamics obey the master equation
\begin{equation}
\frac{d}{dt} p(t) = p(t) Q,
\qquad
p(t) = \pi_0 \exp(Qt).
\end{equation}
This is the main object behind our MJP experiments: once $Q$ and $\pi_0$ are known, one can recover time-dependent state probabilities, simulate new trajectories, and compute several physically meaningful observables.

The inverse problem considered in the main text is difficult because we do not observe a full jump trajectory. Instead, we only see a discrete-time snapshot of the latent process, often on an irregular grid and possibly with observation noise. Between two consecutive observation times, the hidden process may remain in the same state or may undergo multiple unobserved jumps. Estimating $Q$ from such partial information is therefore more challenging than estimating a discrete-time transition matrix from fully observed state sequences.

Several standard quantities used in the MJP literature follow directly from the inferred generator. A stationary distribution $\pi^\star$ satisfies
\begin{equation}
\pi^\star Q = 0,
\qquad
\sum_{i=1}^K \pi^\star_i = 1.
\end{equation}
The dominant relaxation times are determined by the non-zero eigenvalues $\lambda_2, \dots, \lambda_K$ of $Q$ through $-1 / \mathrm{Re}(\lambda_i)$, which characterize how quickly perturbations decay toward stationarity. Mean first-passage times can likewise be obtained by solving linear systems associated with the generator. In addition, exact trajectories can be sampled with the Gillespie stochastic simulation algorithm, which alternates between sampling an exponential waiting time and a discrete next state according to the outgoing rates \cite{gillespie1977exact}.

\subsection{Synthetic Training Data Reused from FIM-MJP}

For the MJP task, we reuse the synthetic training corpus introduced in FIM-MJP \cite{fim_mjp} rather than generating a new synthetic benchmark from scratch. Their data-generation pipeline samples MJPs on state spaces of size 2--6 by first drawing a connected adjacency structure and then sampling the allowed off-diagonal rates from a small family of Beta priors. The initial distribution is chosen either as the stationary distribution of the sampled generator or from a Dirichlet distribution, so the resulting processes cover both equilibrium-like and more generic initializations.

Trajectories are then simulated with the Gillespie algorithm on a time horizon of length $10$ \cite{gillespie1977exact}. To obtain observations, the latent paths are recorded on a base grid with at most 100 time points and then subsampled to produce both regular and irregular observation patterns. Finally, a small amount of label noise is injected by randomly replacing a fraction of observed states; in the version we use, this corruption level is 1\%. The full FIM-MJP corpus contains 45K synthetic processes with 300 sampled paths per process \cite{fim_mjp}.

\subsection{MJP Datasets}

Our MJP experiments cover four qualitatively different settings. Together they test whether a single zero-shot heuristic can handle synthetic non-equilibrium physics, noisy biophysical recordings, and coarse-grained molecular dynamics.

\paragraph{Discrete Flashing Ratchet (DFR).}
The DFR is a synthetic six-state benchmark derived from a Brownian ratchet model \cite{ajdari1992mouvement}. It represents a particle moving on a periodic asymmetric landscape whose potential is switched on and off. In the ``on'' configuration, the asymmetry induces a directional bias; in the ``off'' configuration, the particle diffuses more freely. The switching breaks detailed balance and creates a non-equilibrium steady state with a persistent current, making the DFR a useful benchmark for testing whether inferred generators capture genuinely irreversible dynamics. In our benchmark setup, we follow the dataset used by prior MJP work \cite{neural_mjp,fim_mjp}: 5000 coarse-grained paths observed on irregular grids with 50 observation times per path. Section~\ref{app:dfr_entropy} provides the mathematical details of the DFR construction and the entropy-production calculation.

\paragraph{Ion Channel (IonCh).}
This dataset comes from experimental recordings of the viral potassium channel $\mathrm{Kcv}_{\mathrm{MT325}}$, which exhibits switching between a small number of metastable conductance states \cite{gazzarrini2006chlorella}. The raw signal is continuous and noisy, so the MJP is only obtained after coarse-graining the recordings into discrete channel states. This makes the task harder than the DFR benchmark: besides inferring transition rates, one must cope with measurement noise and the fact that the latent discrete representation is itself model-dependent. The benchmark introduced in the earlier MJP papers compares methods on this three-state switching system \cite{neural_mjp,fim_mjp}.

\paragraph{Alanine Dipeptide (ADP).}
Alanine dipeptide is a canonical molecular-kinetics benchmark because its slow conformational dynamics can be summarized well by the Ramachandran torsion angles $\phi$ and $\psi$ while still exhibiting non-trivial metastability \cite{mironov19}. The observations are continuous molecular descriptors, typically encoded through trigonometric features of these angles, and are then mapped to a small discrete state space for MJP inference. In the benchmark used by previous work, the goal is to recover the coarse-grained switching dynamics among six metastable conformations from all-atom molecular simulation data \cite{mardt17,husic20,neural_mjp,fim_mjp}.

\paragraph{Protein Folding (PFold).}
The PFold benchmark is a simpler two-state molecular system, originating from a Brownian-dynamics model with a bistable potential that mimics transitions between folded and unfolded configurations \cite{mardt17}. After coarse-graining, the task reduces to recovering a two-state generator whose rates determine the switching statistics and stationary occupancy. This dataset is useful because it isolates the basic metastable-switching regime: strong methods should match the correct time scales even when the state space is very small.

For DFR, the ground-truth process is known analytically, so one can directly evaluate parameter recovery. For IonCh, ADP, and PFold, the discrete-state abstraction is obtained from real-valued or molecular trajectories, so evaluation is more indirect and relies on whether the inferred generator reproduces the observed long-time and short-time statistics, as well as the quality of simulated rollouts.

\subsection{Evaluation Metrics}

We use two different types of evaluation for the MJP task. On the synthetic DFR benchmark, where the underlying parameters are known, Table~\ref{tab:DFR} reports the inferred parameters as ratios to the ground truth, \emph{e.g.} $\hat{V}/V$, $\hat{r}/r$, and $\hat{b}/b$. A value of $1$ therefore corresponds to perfect recovery, values above $1$ indicate overestimation, and values below $1$ indicate underestimation.

For the broader rollout comparison across DFR, IonCh, ADP, and PFold, we use the Hellinger distance between empirical state distributions \cite{hellinger1909neue}, following its use as an MJP evaluation metric in FIM-MJP \cite{fim_mjp}. Intuitively, it measures how similar the predicted path distribution is to the true path distribution over time. This metric is especially useful for the real-world datasets, where the true latent generator is not known and parameter error therefore cannot be computed directly. Given two discrete distributions $P=(p_1,\dots,p_K)$ and $Q=(q_1,\dots,q_K)$, the Hellinger distance is
\begin{equation}
H(P,Q) = \frac{1}{\sqrt{2}}
\sqrt{
\sum_{i=1}^{K}
\left(\sqrt{p_i}-\sqrt{q_i}\right)^2
}.
\end{equation}
In our setting the distributions are not available in closed form, so we approximate them from sampled or observed paths. At each observation time $\tau_\ell$, we form normalized histograms $h^{\mathrm{tar}}_\ell$ and $h^{\mathrm{mod}}_\ell$ over the discrete states using the target process and the model-generated process, respectively. We then compute the local discrepancy $H(h^{\mathrm{tar}}_\ell, h^{\mathrm{mod}}_\ell)$ and average it over the observation grid:
\begin{equation}
d_{\mathrm{Hell}}
=
\frac{1}{L}
\sum_{\ell=1}^{L}
H\!\left(h^{\mathrm{tar}}_\ell, h^{\mathrm{mod}}_\ell\right).
\end{equation}
Lower values indicate that the simulated process matches the empirical occupancy statistics more closely over time. Following FIM-MJP \cite{fim_mjp}, we report means and standard deviations over 100 repeated histogram comparisons.

\subsection{DFR and Entropy Production}
\label{app:dfr_entropy}

The discrete flashing ratchet (DFR) is a six-state non-equilibrium MJP obtained by coarse-graining a Brownian particle moving in an asymmetric periodic potential that is switched on and off \cite{ajdari1992mouvement}. As in the FIM-MJP setup \cite{fim_mjp}, the ``on'' states correspond to motion in the asymmetric potential and the ``off'' states to freer diffusion. The transition rates are parameterized by the potential strength $V$, the switching rate $r$, and a background diffusion rate $b$. For the within-mode dynamics one writes
\begin{equation}
f_{ij}^{\mathrm{on}} = \exp\!\left(-\frac{V}{2}(j-i)\right),
\qquad
f_{ij}^{\mathrm{off}} = b,
\end{equation}
for positions $i,j \in \{1,2,3\}$ within the on and off sectors, respectively, while the on/off switching itself occurs at rate $r$ between corresponding positions. Writing the six states as $\{1_{\mathrm{on}},2_{\mathrm{on}},3_{\mathrm{on}},1_{\mathrm{off}},2_{\mathrm{off}},3_{\mathrm{off}}\}$, the exact generator is equivalently specified by its nonzero entries:
\begin{align}
Q_{i_{\mathrm{on}},j_{\mathrm{on}}} &= \exp\!\left(-\frac{V}{2}(j-i)\right),
&& i \neq j,\; i,j \in \{1,2,3\},\\
Q_{i_{\mathrm{off}},j_{\mathrm{off}}} &= b,
&& i \neq j,\; i,j \in \{1,2,3\},\\
Q_{i_{\mathrm{on}},i_{\mathrm{off}}} &= Q_{i_{\mathrm{off}},i_{\mathrm{on}}} = r,
&& i \in \{1,2,3\},\\
Q_{\alpha,\beta} &= 0,
&& \text{otherwise},\\
Q_{\alpha,\alpha} &= - \sum_{\beta \neq \alpha} Q_{\alpha,\beta}.
\end{align}
Setting $(V,r,b)=(1,1,1)$ recovers the ground-truth matrix used in the benchmark tables up to rounding. Intuitively, increasing $V$ increases the asymmetry of the landscape, $b$ sets the scale of unbiased motion when the potential is off, and $r$ controls how often the system alternates between these two regimes.

This model is useful because the asymmetry and random switching create a stationary current, so the process is out of equilibrium. In other words, the DFR does not merely have a stationary distribution; it also exhibits irreversible probability flow around the state space. This makes it a natural benchmark for testing whether an inferred generator recovers not only occupancies, but also the directional structure of the dynamics.

For finite-state MJPs, the instantaneous total entropy production rate can be written in closed form from the generator and the time-dependent probabilities \cite{Seifert_2012}:
\begin{equation}
\dot{S}_{\mathrm{tot}}(t)
=
\frac{1}{2}
\sum_{i,j}
\left[
p_i(t) Q_{ij} - p_j(t) Q_{ji}
\right]
\log
\frac{p_i(t) Q_{ij}}{p_j(t) Q_{ji}}.
\end{equation}
The total entropy production over a time horizon $[0,T]$ is then
\begin{equation}
\Sigma_T = \int_0^T \dot{S}_{\mathrm{tot}}(t)\, dt.
\end{equation}
At equilibrium, detailed balance makes this quantity vanish. For the DFR, by contrast, the ratchet current leads to positive entropy production, which is why it is a meaningful downstream observable in our experiments. Once a method has inferred $Q$ and $\pi_0$, it can solve the master equation for $p(t)$ and evaluate the expression above without any extra training. The entropy-production plot in the main text should therefore be read as a derived scientific quantity that tests whether the inferred MJP captures the correct irreversible dynamics.

\section{Additional Imputation Dataset Details}
\label{app:imputation_details}

Our imputation experiments follow the same benchmark datasets and preprocessing conventions used by the FIM imputation work \cite{seifner2025zeroshot}. We group the datasets into point-wise missingness tasks, where individual values are masked independently, and window-missing tasks, where a contiguous middle segment is removed.

\subsection{Point-Wise Missing Datasets}

Eight datasets use 50\% random point-wise missing patterns. Following \cite{seifner2025zeroshot}, the \dataset{GuangZhou} traffic-speed dataset (214 channels, 500 observations) and the \dataset{Solar} solar-power dataset (137 channels, 52{,}560 observations) come from the benchmark preprocessing used by BayOTIDE \cite{fang2023bayotide}. The remaining six datasets come from TSI-Bench \cite{du2024tsi}: \dataset{Beijing} air quality (132 channels, 1{,}458 observations), \dataset{Italy} air quality (13 channels, 774 observations), \dataset{Electricity} consumption (370 channels, 1{,}457 observations), \dataset{ETT\_h1} electricity transformer temperature / load (7 channels, 358 observations), \dataset{PeMS} road occupancy (862 channels, 727 observations), and \dataset{Pedestrian} activity in Australia (1 channel, 3{,}633 observations). As in the benchmark protocol, the datasets are split into train, validation, and test subsets, and 50\% of each subset is randomly masked and used for evaluation.

\subsection{Window-Missing Datasets}

For contiguous-gap imputation, we use the Motion Capture dataset of human walking motions \cite{wang2007gaussian}, as preprocessed by \cite{yildiz2019ode2vae}, following the benchmark setup of \cite{heinonen2018learning,seifner2025zeroshot}. The dataset contains 43 human motion trajectories, each with 50 channels and 100 observations. Roughly 20\% of the middle portion of each time series is removed, creating a temporally missing window that must be reconstructed from the surrounding context. We report two variants of this benchmark: \textit{MC (PCA)}, where the sequences are projected to a lower-dimensional PCA representation before imputation, and \textit{MC (No PCA)}, where imputation is performed directly in the original data space.

\section{Discovered Algorithms}
\label{app:algorithms}

\begin{pseudocodebox}{algpseud:tpp}{\ours heuristic for Point Processes}
\textbf{Input:} context event sequences, target prefixes, number of marks $K$.

\textbf{Context statistics.}
\begin{enumerate}
\item Aggregate all context inter-event gaps and all mark transitions $a \to b$.
\item Estimate a robust global gap statistic from all observed gaps.
\item For each previous mark $a$, estimate the average next-event gap after observing $a$.
\item Build a smoothed transition table $P(b \mid a)$ from the context sequences.
\item Estimate the global majority mark as a fallback.
\end{enumerate}

\textbf{Prediction for one target prefix.}
\begin{enumerate}
\item If the prefix is empty, predict the global gap and the global majority mark.
\item Otherwise, let $a$ be the last observed mark and compute a recency-weighted average of the most recent gaps in the prefix.
\item Predict the next gap by mixing this local recent-gap estimate with the context-level average gap associated with mark $a$; clamp extreme values.
\item Predict the next time as the last observed time plus the predicted gap.
\item Predict the next mark by combining local counts of transitions out of $a$ in the prefix, mark frequencies within the prefix, and the smoothed context transition row $P(\cdot \mid a)$.
\item Return the predicted next time and the highest-scoring next mark.
\end{enumerate}
\end{pseudocodebox}

\begin{pythoncode}[label=alg:tpp]{\ours algorithm for Point Processes}
"""
Zero-shot MTPP next-event prediction heuristic (evolvable block).

The function predict_next_events is called by the evaluator with batched
target histories and a context pool; it must return predicted next times and marks.
"""

import numpy as np

def predict_next_events(
    target_history_times,
    target_history_marks,
    context_times,
    context_marks,
    num_classes,
):
    b = len(target_history_times)
    pt = np.zeros(b, np.float64)
    pm = np.zeros(b, np.int64)

    # Context: robust gap, majority mark, transition matrix, and per-mark mean delta
    deltas = []
    counts = np.zeros(num_classes, np.int64)
    trans = np.zeros((num_classes, num_classes), np.float64)
    g_sum = np.zeros(num_classes, np.float64)
    g_cnt = np.zeros(num_classes, np.int64)

    for ta, ma in zip(context_times, context_marks):
        t = np.asarray(ta, np.float64)
        m = np.asarray(ma, np.int64)
        if t.size >= 2:
            d = np.diff(t)
            deltas.extend(d.tolist())
            # accumulate mean delta conditioned on the previous mark
            for prev_mark, dd in zip(m[:-1], d):
                if 0 <= prev_mark < num_classes:
                    g_sum[int(prev_mark)] += float(dd)
                    g_cnt[int(prev_mark)] += 1
        if m.size:
            counts += np.bincount(m, minlength=num_classes)
        if m.size >= 2:
            a = m[:-1]
            b2 = m[1:]
            for x, y in zip(a, b2):
                if 0 <= x < num_classes and 0 <= y < num_classes:
                    trans[int(x), int(y)] += 1.0

    ggap = float(np.median(deltas)) if deltas else 1.0
    gmaj = int(np.argmax(counts)) if counts.sum() > 0 else 0

    # Prepare a smoothed global transition baseline to avoid zero rows
    trans_baseline = trans + 1.0  # Laplace-like smoothing
    trans_baseline /= trans_baseline.sum(axis=1, keepdims=True) + 1e-12

    # Per-mark global mean delta (fallback)
    g_mean_delta = np.full(num_classes, ggap, dtype=np.float64)
    mask = g_cnt > 0
    g_mean_delta[mask] = g_sum[mask] / g_cnt[mask]

    for i in range(b):
        t = np.asarray(target_history_times[i], np.float64)
        m = np.asarray(target_history_marks[i], np.int64)
        n = t.size

        if n == 0:
            pt[i] = ggap
            pm[i] = gmaj
            continue

        last_t = float(t[-1])
        last_m = int(m[-1]) if m.size else -1

        # Time prediction: recency-weighted gaps with mark-conditioned fallback and clamping
        if n >= 2:
            d = np.diff(t)
            recent = d[-5:]  # up to last 5 intervals
            Lr = recent.size
            # exponential recency weights (more recent -> higher)
            exps = np.exp(np.linspace(-1.0, 0.0, max(1, Lr)))
            exps = exps[-Lr:]
            exps = exps / (exps.sum() + 1e-12)
            rec_est = float(np.dot(recent, exps))
            if 0 <= last_m < num_classes and g_cnt[last_m] > 0:
                cond_mean = g_mean_delta[last_m]
                base = 0.6 * rec_est + 0.4 * cond_mean
            else:
                base = 0.75 * rec_est + 0.25 * ggap
            base = float(max(base, 1e-6))
            # cap extreme predictions to a reasonable multiple
            cap = 20.0 * max(rec_est if rec_est > 0 else ggap, ggap)
            base = min(base, cap)
            pt[i] = last_t + base
        else:
            # Single time in history -> use per-mark mean delta if available, else global median
            if 0 <= last_m < num_classes and g_cnt[last_m] > 0:
                delta = float(g_mean_delta[last_m])
            else:
                delta = ggap
            delta = max(delta, 1e-6)
            pt[i] = last_t + delta

        # Mark prediction: combine local transition counts, local frequency, and global smoothed transitions
        scores = np.zeros(num_classes, dtype=np.float64)

        # 1) local transitions conditioned on last mark
        if m.size >= 2:
            prev = m[:-1]
            nxt = m[1:]
            mask_prev = prev == last_m
            if np.any(mask_prev):
                counts_local_next = np.bincount(nxt[mask_prev], minlength=num_classes).astype(np.float64)
                scores += 1.0 * counts_local_next

        # 2) local frequency in the sequence (helps for short histories)
        hist_counts = np.bincount(m, minlength=num_classes).astype(np.float64)
        scores += 0.25 * hist_counts

        # 3) global smoothed transition probabilities from last mark (as soft prior)
        if 0 <= last_m < num_classes:
            scores += 2.0 * trans_baseline[last_m]
        else:
            # if last mark invalid, fallback to global popularity
            scores += 0.5 * (counts + 1.0)

        # If everything is zero (very unlikely), fallback to global majority
        if scores.sum() <= 0:
            pm[i] = gmaj
        else:
            pm[i] = int(np.argmax(scores))

    return pt, pm
\end{pythoncode}
\FloatBarrier

\begin{pseudocodebox}{algpseud:tpp_synthetic}{\ourssynth heuristic for Point Processes}
\textbf{Input:} context event sequences, target prefixes, number of marks $K$.

\textbf{Context statistics.}
\begin{enumerate}
\item Aggregate all inter-event gaps from the context sequences.
\item Build first-order transition counts $a \to b$ between consecutive marks.
\item Build second-order transition counts $(a,b) \to c$ for consecutive mark pairs.
\item For each previous mark $a$, estimate a typical outgoing gap; for each next mark $b$, estimate a typical incoming gap.
\item For each directed edge $a \to b$, estimate an edge-specific typical gap whenever enough examples exist.
\item Estimate a global fallback gap and a global majority mark.
\end{enumerate}

\textbf{Prediction for one target prefix.}
\begin{enumerate}
\item Predict the next mark using a reliability hierarchy:
\begin{enumerate}
\item if the last two marks form a previously observed pair, use the second-order transition table;
\item otherwise, if the last mark has outgoing transitions in the context data, use the first-order transition table;
\item otherwise, fall back to the global majority mark.
\end{enumerate}
\item Predict the next gap using an edge-specific timing estimate for the predicted transition whenever available.
\item If no edge-specific estimate is available, combine the typical outgoing gap of the last mark with the typical incoming gap of the predicted mark.
\item Mix this context-level timing estimate with the recent local gaps from the target prefix.
\item Predict the next time as the last observed time plus the resulting gap.
\item Return the predicted next time and mark.
\end{enumerate}
\end{pseudocodebox}

\begin{pythoncode}[label=alg:tpp_synthetic]{\ourssynth algorithm for Point Processes}
"""
Zero-shot MTPP next-event prediction heuristic (evolvable block).

The function predict_next_events is called by the evaluator with batched
target histories and a context pool; it must return predicted next times and marks.
"""

import numpy as np

def predict_next_events(
    target_history_times,
    target_history_marks,
    context_times,
    context_marks,
    num_classes,
):
    b = len(target_history_times)
    # Context statistics: global, 1st- and 2nd-order mark transitions, and mark-conditional deltas
    trans = np.zeros((num_classes, num_classes), dtype=np.int64)
    per_prev = [[] for _ in range(num_classes)]
    per_next = [[] for _ in range(num_classes)]
    pair = {}
    edge_dt = {}
    all_dt, all_m = [], []
    for t, m in zip(context_times, context_marks):
        Lm = len(m)
        if len(t) > 1:
            dt = np.diff(t)
            all_dt.extend(dt.tolist())
            L = min(len(dt), Lm - 1) if Lm > 0 else 0
            if L > 0:
                prev = m[:L]
                nxt = m[1:1 + L]
                for j in range(L):
                    d = float(dt[j])
                    a = int(prev[j])
                    bmark = int(nxt[j])
                    per_prev[a].append(d)
                    per_next[bmark].append(d)
                    trans[a, bmark] += 1
                    edge_dt.setdefault((a, bmark), []).append(d)
            if Lm > 2:
                for j in range(Lm - 2):
                    key = (int(m[j]), int(m[j + 1]))
                    if key in pair:
                        pair[key][m[j + 2]] += 1
                    else:
                        cnt = np.zeros(num_classes, dtype=np.int64)
                        cnt[m[j + 2]] = 1
                        pair[key] = cnt
        if Lm > 0:
            all_m.extend(m)
    gmd = float(np.median(all_dt)) if all_dt else 1.0
    gmm = int(np.bincount(all_m).argmax()) if all_m else 0
    pmd = np.full(num_classes, gmd, dtype=float)
    nmd = np.full(num_classes, gmd, dtype=float)
    for c in range(num_classes):
        if per_prev[c]:
            pmd[c] = float(np.median(per_prev[c]))
        if per_next[c]:
            nmd[c] = float(np.median(per_next[c]))
    next_by_prev = (trans + 1).argmax(axis=1)  # Laplace smoothing
    row_counts = trans.sum(axis=1)
    # per-transition median delta (edge-specific)
    edge_md = {k: float(np.median(v)) for k, v in edge_dt.items()} if edge_dt else {}

    T = np.zeros(b, dtype=np.float64)
    M = np.zeros(b, dtype=np.int64)
    for i in range(b):
        th = target_history_times[i]
        mh = target_history_marks[i]
        ln = len(mh)
        lm = int(mh[-1]) if ln > 0 else -1

        # Predict mark with reliability checks
        if ln >= 2:
            key = (int(mh[-2]), lm)
            if key in pair and pair[key].sum() > 0:
                nm = int(pair[key].argmax())
            elif 0 <= lm < num_classes and row_counts[lm] > 0:
                nm = int(next_by_prev[lm])
            else:
                nm = gmm
        elif ln == 1:
            if 0 <= lm < num_classes and row_counts[lm] > 0:
                nm = int(next_by_prev[lm])
            else:
                nm = gmm
        else:
            nm = gmm
        M[i] = int(nm)

        # Predict time using local recent median and edge-specific/context medians
        if len(th) > 0:
            base = float(th[-1])
            # reference delta from edge if available, else combine per-prev and per-next, else global
            if ln > 0 and 0 <= lm < num_classes and 0 <= nm < num_classes:
                ek = (lm, nm)
                if ek in edge_md:
                    dref = float(edge_md[ek])
                else:
                    dref = 0.5 * (pmd[lm] + nmd[nm])
            else:
                dref = gmd

            if len(th) > 1:
                k = min(5, len(th))
                dloc = float(np.median(np.diff(th[-k:])))
                alpha = 0.65 if len(th) >= 3 else 0.5
                d = alpha * dloc + (1.0 - alpha) * dref
            else:
                d = dref
            T[i] = base + d
        else:
            T[i] = gmd
    return T, M
\end{pythoncode}
\FloatBarrier

\begin{pseudocodebox}{algpseud:mjp}{\ourssynth heuristic for Markov Jump Processes}
\textbf{Input:} partially observed trajectories, observation times, sequence lengths, number of states $K$.

\textbf{Initial distribution.}
\begin{enumerate}
\item Count the first observed state of each trajectory.
\item Add smoothing, and optionally add a smaller contribution from second observations to reduce sensitivity to noise.
\item Normalize the counts to obtain the initial-state distribution.
\end{enumerate}

\textbf{Characteristic time scale.}
\begin{enumerate}
\item Collect valid time differences between consecutive observations.
\item Use a robust typical interval, such as the median positive difference.
\end{enumerate}

\textbf{Rate matrix estimation.}
\begin{enumerate}
\item For each observed interval spent in state $i$, add its duration to the exposure time of state $i$, while capping unusually long intervals.
\item Count observed exits from each state, ignoring implausibly fast changes that are likely due to noise.
\item For each valid transition $i \to j$, accumulate a destination count, giving larger weight to transitions observed over shorter intervals.
\item Estimate each state's exit hazard as a smoothed ratio of exits to exposure time, then clip it to a reasonable range.
\item Normalize destination counts from each state to obtain off-diagonal transition probabilities.
\item Form the rate matrix by multiplying each state's exit hazard with its destination distribution, then set the diagonal so that each row sums to zero.
\end{enumerate}
\end{pseudocodebox}

\begin{pythoncode}[label=alg:mjp_synthetic]{\ourssynth algorithm for Markov Jump Processes}
"""
Zero-shot MJP parameter estimation heuristic (evolvable block).
"""

import numpy as np

def estimate_mjp_parameters(
    observation_grid: np.ndarray,
    observation_values: np.ndarray,
    seq_lengths: np.ndarray,
    n_states: int,
):
    """
    Predict the global rate matrix and initial distribution for one MJP sample.
    Inputs are for a single sample: num_paths trajectories, each with up to max_seq_len observations.

    Args:
        observation_grid: (num_paths, max_seq_len) array of observation times.
        observation_values: (num_paths, max_seq_len) array of observed discrete states (0 to n_states-1).
        seq_lengths: (num_paths,) array indicating the number of valid observations per path.
        n_states: The number of states for this MJP.

    Returns:
        Tuple of:
        - pred_rate_matrix: (n_states, n_states) array. Off-diagonals are >= 0.
        - pred_init_dist: (n_states,) array of probabilities summing to 1.
    """
    batch_size, max_len = observation_grid.shape

    # -------------------------------------------------------------------------
    # 1. Estimate Initial Distribution globally (Vectorized)
    # -------------------------------------------------------------------------
    valid_paths = seq_lengths > 0
    if not np.any(valid_paths):
        # Fallback if entirely empty sequence batch
        return np.zeros((n_states, n_states)), np.ones(n_states) / n_states

    # Grab first observations with uniform prior
    first_obs = observation_values[valid_paths, 0].astype(int)
    counts = np.bincount(first_obs, minlength=n_states).astype(np.float64) + 1.0

    # Grab second observations (if available) with reduced weight to smooth noise
    valid_len2 = seq_lengths > 1
    if np.any(valid_len2):
        second_obs = observation_values[valid_len2, 1].astype(int)
        counts += np.bincount(second_obs, minlength=n_states) * 0.5

    global_init_dist = counts / np.sum(counts)

    # -------------------------------------------------------------------------
    # 2. Extract Valid Intervals & Estimate Time Scale (Vectorized)
    # -------------------------------------------------------------------------
    # Create a boolean mask of valid transitions: (batch, max_len - 1)
    mask = np.arange(max_len - 1)[None, :] < (seq_lengths[:, None] - 1)
    
    diffs = np.diff(observation_grid, axis=1)
    valid_dts = diffs[mask]
    
    pos_dts = valid_dts[valid_dts > 0]
    typical_dt = float(np.median(pos_dts)) if pos_dts.size > 0 else 1.0
    typical_dt = max(1e-8, typical_dt)

    # -------------------------------------------------------------------------
    # 3. Accumulate State Exits and Exposures
    # -------------------------------------------------------------------------
    # Extract flattened arrays of valid states and times
    curr_states = observation_values[:, :-1][mask].astype(int)
    next_states = observation_values[:, 1:][mask].astype(int)
    dts = valid_dts

    # Extract the last valid state of each path for right-censoring exposure
    last_idx = np.maximum(0, seq_lengths - 1)
    last_states = observation_values[np.arange(batch_size), last_idx][valid_paths].astype(int)

    # Compute total exposure time per state (capped to prefer small/medium intervals)
    capped_dts = np.minimum(dts, 2.0 * typical_dt)
    time_small = np.bincount(curr_states, weights=capped_dts, minlength=n_states)
    time_small += np.bincount(last_states, weights=np.full(last_states.shape, 0.5 * typical_dt), minlength=n_states)

    # Filter out transitions that are too fast (likely noise)
    valid_change = (curr_states != next_states) & (dts > 0.02 * typical_dt)
    change_small = np.bincount(curr_states[valid_change], minlength=n_states)

    # Accumulate destination targets, weighted by recency
    dest_counts = np.zeros((n_states, n_states), dtype=np.float64)
    weights = np.exp(-dts[valid_change] / typical_dt)
    np.add.at(dest_counts, (curr_states[valid_change], next_states[valid_change]), weights)

    # -------------------------------------------------------------------------
    # 4. Assemble Rate Matrix
    # -------------------------------------------------------------------------
    # Estimate per-state exit rates (hazard)
    alpha = (change_small + 0.1) / (time_small + 0.25 * typical_dt)
    alpha = np.clip(alpha, 1e-8 / typical_dt, 10.0 / typical_dt)

    # Compute destination transition probabilities
    P_off = dest_counts + (0.05 / max(1, n_states - 1))
    np.fill_diagonal(P_off, 0.0) # Ensure no self-loops exist in the probability matrix
    
    row_sums = P_off.sum(axis=1, keepdims=True)
    row_sums[row_sums == 0] = 1.0  # Prevent division by zero if a state has no outward paths
    P_off /= row_sums

    # Construct the final Q-matrix: Q = diag(alpha) * P_off
    global_rate_matrix = alpha[:, None] * P_off
    
    # Strictly enforce Q-matrix eigenvalue constraint: rows must sum to exactly zero
    np.fill_diagonal(global_rate_matrix, 0.0)
    global_rate_matrix[np.diag_indices(n_states)] = -global_rate_matrix.sum(axis=1)

    return global_rate_matrix, global_init_dist
\end{pythoncode}
\FloatBarrier

\begin{pseudocodebox}{algpseud:imputation}{\ours heuristic for Time Series Imputation}
\textbf{Input:} partially observed time series $x_{1:T}$ with timestamps.

\textbf{Process each dimension independently.}
\begin{enumerate}
\item Detect contiguous blocks of missing values.
\item If the series is entirely missing, fill it with a simple constant fallback.
\item For each missing block:
\begin{enumerate}
\item If the gap is short, leave it for the interpolation fallback.
\item If the gap is long enough and sufficient observed context exists before it, extract the observed window immediately preceding the gap.
\item Search earlier in the same series for candidate windows whose preceding context is fully observed and has the same length.
\item Score each candidate by how well its preceding context matches the current pre-gap context.
\item Copy the continuation that follows the best matching window into the gap.
\item Apply a level shift so the copied pattern connects smoothly to the value right before the gap.
\end{enumerate}
\item After motif retrieval has been attempted for all long gaps, fill any remaining missing positions by time-aware linear interpolation.
\end{enumerate}
\end{pseudocodebox}

\begin{pythoncode}[label=alg:imputation]{\ours algorithm for time series imputation}
"""
Time series imputation heuristic (evolvable block).

Data contains randomly masked point-wise missing patterns.
The function receives `observation_values` (T, D) — batched across D channels within one dataset —
with NaN at missing/holdout positions, `observation_times` (T,) per time step, and
`prediction_mask` (T, D) True at holdout positions we must predict.
"""

import numpy as np
from numpy.lib.stride_tricks import sliding_window_view

def _impute_1d(out: np.ndarray, times: np.ndarray) -> None:
    """
    Time-aware hybrid imputation for a single 1D series.
    Uses vectorized motif retrieval for large gaps and linear interpolation for small gaps.
    """
    large_gap_threshold = 4
    context_size = 8
    
    valid = np.isfinite(out)
    
    if not np.any(valid):
        out[:] = 0.0
        return
    if np.all(valid):
        return

    # 1. Identify contiguous missing blocks (vectorized)
    is_missing = ~valid
    padded = np.concatenate(([False], is_missing, [False]))
    starts = np.where(padded[1:] & ~padded[:-1])[0]
    ends = np.where(~padded[1:] & padded[:-1])[0]
    
    for s, e in zip(starts, ends):
        gap_len = e - s
        
        if gap_len >= large_gap_threshold and s >= context_size:
            context = out[s - context_size : s]
            
            if np.all(np.isfinite(context)):
                req_len = context_size + gap_len
                search_end = s - req_len + 1
                
                if search_end > 0:
                    # Search space is strictly before the current gap
                    search_space_valid = valid[:search_end + req_len - 1]
                    
                    # 2. Vectorized identification of valid windows
                    windows_valid = sliding_window_view(search_space_valid, req_len).all(axis=1)
                    valid_indices = np.where(windows_valid)[0]
                    
                    if len(valid_indices) > 0:
                        # 3. Vectorized extraction of candidate contexts
                        search_space_vals = out[:search_end + context_size - 1]
                        all_cand_contexts = sliding_window_view(search_space_vals, context_size)
                        
                        valid_cand_contexts = all_cand_contexts[valid_indices]
                        
                        # 4. Vectorized distance calculation
                        distances = np.sum((valid_cand_contexts - context)**2, axis=1)
                        
                        best_idx_in_valid = np.argmin(distances)
                        best_match_idx = valid_indices[best_idx_in_valid]
                        
                        # 5. Inject the retrieved motif
                        retrieved_pattern = out[best_match_idx + context_size : best_match_idx + context_size + gap_len]
                        
                        # Apply level shift
                        shift = context[-1] - out[best_match_idx + context_size - 1]
                        out[s : e] = retrieved_pattern + shift
                        
                        # Mark newly imputed block as valid
                        valid[s : e] = True
                    
    # 6. Fallback: time-aware linear interpolation
    valid = np.isfinite(out)
    if not np.all(valid):
        valid_times = times[valid]
        valid_vals = out[valid]
        out[:] = np.interp(times, valid_times, valid_vals)


def impute(
    observation_values: np.ndarray,
    observation_times: np.ndarray,
    prediction_mask: np.ndarray,
) -> np.ndarray:
    """
    Impute values at positions where prediction_mask is True.
    observation_values: (T, D) array, NaN at unobserved and holdout.
    observation_times: (T,) timestamps for each time step.
    prediction_mask: (T, D) True at positions we must predict.

    Returns:
        Full array same shape (T, D), with all NaNs filled (imputed).
    """
    out = np.array(observation_values, copy=True, dtype=np.float64)
    times = np.asarray(observation_times, dtype=np.float64)

    if out.ndim == 1:
        _impute_1d(out, times)
        return out

    T, D = out.shape
    if T == 0 or D == 0:
        return out

    for d in range(D):
        _impute_1d(out[:, d], times)

    return out
\end{pythoncode}
\FloatBarrier

\section{Hyperparameters and System Prompts}
\label{app:hyperparams}

All experiments use OpenEvolve \cite{sharma2025openevolve}. Across tasks, we used the same LLM backbone and the same core evolutionary setup: a weighted ensemble with \texttt{gpt-5-mini-2025-08-07} (weight $0.8$) and \texttt{gpt-5-2025-08-07} (weight $0.2$), temperature $0.7$, \texttt{max\_tokens}=16{,}000, model timeout $120$s, diff-based evolution, \texttt{max\_code\_length}=20{,}000, archive size $20$, population size $50$, and MAP-Elites with $3$ islands. All runs used \texttt{elite\_selection\_ratio}=0.2 and \texttt{similarity\_threshold}=0.99; \texttt{exploitation\_ratio} was $0.7$. The evaluator used \texttt{parallel\_evaluations}=5 and \texttt{cascade\_thresholds}=\{0.0\} in all three tasks. Evaluator timeouts were $60$s for point processes and MJP, and $3600$s for imputation.

\paragraph{Point processes.}
The point-process task optimized a zero-shot heuristic for next-event time and mark prediction. The score was
\[
\text{score} = \text{accuracy} - \lambda \cdot \text{RMSE}_{\Delta t}, \qquad \lambda = 1.
\]
The exact system prompt was:
\begin{Verbatim}[breaklines=true,breakanywhere=true,fontsize=\small]
You are an expert statistician and algorithm designer for temporal point processes and marked event sequences.
Your task is to improve a zero-shot Python heuristic that predicts the next event time and mark (type) given
a batch of sequence histories and a context pool of sequences.

CONSTRAINTS (strict):
- Do NOT use PyTorch, TensorFlow, JAX, or any other heavy ML framework.
- Use only standard library Python and NumPy.
- All code must stay inside the EVOLVE-BLOCK and implement the function predict_next_events with this exact signature:
  predict_next_events(target_history_times, target_history_marks, context_times, context_marks, num_classes)
- The function must return a tuple (predicted_next_times, predicted_next_marks), both 1D NumPy arrays of length batch_size.
\end{Verbatim}

\paragraph{Markov jump processes.}
For MJP, the evolved program estimated the initial distribution and continuous-time rate matrix from discretely observed trajectories. The score combined the cross-entropy of the initial distribution and the RMSE of the off-diagonal rate entries,
\[
\text{score} = -\text{CE} - \text{RMSE}_Q.
\]
The evaluator used a cascade with threshold $0.0$ before running the full metric computation. The exact system prompt was:
\begin{Verbatim}[breaklines=true,breakanywhere=true,fontsize=\small]
You are an expert in stochastic processes, Markov Jump Processes (MJP), and numerical algorithms.
Your task is to write a pure Python/NumPy heuristic that estimates the continuous-time
transition rate matrix (intensity matrix Q) and the initial probability distribution.

CRITICAL DOMAIN KNOWLEDGE:
1. The data consists of *discrete snapshots* (recordings at specific times), NOT exact jump times. You do not see every state transition. Between `observation_grid[i]` and `observation_grid[i+1]`, zero, one, or multiple hidden jumps could have occurred.
2. The data is noisy. The `observation_values` might contain measurement errors.

Return NO text outside the python code block. Do NOT use torch. Use only numpy.
\end{Verbatim}

\paragraph{Imputation.}
For imputation, the evolved function filled held-out entries in reduced time series, and the score was the negative mean MAE across datasets,
\[
\text{score} = -\text{mean MAE}.
\]
The config used the datasets \textsc{Beijing}, \textsc{Italy}, \textsc{GuangZhou}, \textsc{PeMS}, \textsc{Pedestrian}, \textsc{Solar}, \textsc{ETT\_h1}, \textsc{Electricity}, \textsc{MotionCapture\_PCA}, and \textsc{MotionCapture\_NoPCA}. The standard evaluation holdout fraction was $10\%$ with seed $42$ and a global cap of $1000$ timesteps; for the two MotionCapture datasets, the holdout fraction was increased to $20\%$ and averaged over $10$ folds. The exact system prompt was:
\begin{Verbatim}[breaklines=true,breakanywhere=true,fontsize=\small]
You are an expert in time series analysis, signal processing, and advanced imputation algorithms.
Your task is to improve a zero-shot Python heuristic that predicts values at positions indicated by a mask.
Your algorithm should capture both complex oscillatory datasets but also work well with simple datasets.
The data is REDUCED: only originally-observed positions are kept. The only "missing" values are the eval holdout (NaN). Note that these missing values can be windows of missing values or random missing values.
The function receives observation_values (NaN at holdout), observation_times (exact timestamps for each position), and prediction_mask (True at positions to predict).
Note that it could also be that all elements for one channel are NaN so you need to handle this case as well.
You are allowed to use numpy. If possible, try to use vectorized operations for efficiency.

ALGORITHMIC DIVERSITY & INSPIRATION:
CONSTRAINTS (strict):
- All code must stay inside the EVOLVE-BLOCK and implement the function impute with this exact signature:
  impute(observation_values, observation_times, prediction_mask)
- observation_values: reduced 1D array, with NaN at holdout positions.
- observation_times: 1D array of timestamps for each position (same length as observation_values).
- prediction_mask: boolean array, True at positions we must predict.
- The function must return an array of the same shape as observation_values with all NaNs filled (imputed).
- Lower MAE at the predicted positions is better.
\end{Verbatim}


\newpage
\section*{NeurIPS Paper Checklist}

\begin{enumerate}

\item {\bf Claims}
    \item[] Question: Do the main claims made in the abstract and introduction accurately reflect the paper's contributions and scope?
    \item[] Answer: \answerYes{}
    \item[] Justification: The abstract and introduction state the main claims conservatively and these are supported by the empirical sections across the three tasks, together with the appendices. In particular, the scope of the contribution as interpretable zero-shot inference functions discovered by LLM-guided evolution is described in the Introduction and evaluated in the point-process, Markov-jump-process, and time-series-imputation sections.
    \item[] Guidelines:
    \begin{itemize}
        \item The answer \answerNA{} means that the abstract and introduction do not include the claims made in the paper.
        \item The abstract and/or introduction should clearly state the claims made, including the contributions made in the paper and important assumptions and limitations. A \answerNo{} or \answerNA{} answer to this question will not be perceived well by the reviewers. 
        \item The claims made should match theoretical and experimental results, and reflect how much the results can be expected to generalize to other settings. 
        \item It is fine to include aspirational goals as motivation as long as it is clear that these goals are not attained by the paper. 
    \end{itemize}

\item {\bf Limitations}
    \item[] Question: Does the paper discuss the limitations of the work performed by the authors?
    \item[] Answer: \answerYes{}
    \item[] Justification: The paper includes a dedicated Limitations section discussing weaker performance on some tasks, uncertainty about the novelty of discovered heuristics, determinism of the learned programs, and uncertainty about generalization beyond the tested settings.
    \item[] Guidelines:
    \begin{itemize}
        \item The answer \answerNA{} means that the paper has no limitation while the answer \answerNo{} means that the paper has limitations, but those are not discussed in the paper. 
        \item The authors are encouraged to create a separate ``Limitations'' section in their paper.
        \item The paper should point out any strong assumptions and how robust the results are to violations of these assumptions (e.g., independence assumptions, noiseless settings, model well-specification, asymptotic approximations only holding locally). The authors should reflect on how these assumptions might be violated in practice and what the implications would be.
        \item The authors should reflect on the scope of the claims made, e.g., if the approach was only tested on a few datasets or with a few runs. In general, empirical results often depend on implicit assumptions, which should be articulated.
        \item The authors should reflect on the factors that influence the performance of the approach. For example, a facial recognition algorithm may perform poorly when image resolution is low or images are taken in low lighting. Or a speech-to-text system might not be used reliably to provide closed captions for online lectures because it fails to handle technical jargon.
        \item The authors should discuss the computational efficiency of the proposed algorithms and how they scale with dataset size.
        \item If applicable, the authors should discuss possible limitations of their approach to address problems of privacy and fairness.
        \item While the authors might fear that complete honesty about limitations might be used by reviewers as grounds for rejection, a worse outcome might be that reviewers discover limitations that aren't acknowledged in the paper. The authors should use their best judgment and recognize that individual actions in favor of transparency play an important role in developing norms that preserve the integrity of the community. Reviewers will be specifically instructed to not penalize honesty concerning limitations.
    \end{itemize}

\item {\bf Theory assumptions and proofs}
    \item[] Question: For each theoretical result, does the paper provide the full set of assumptions and a complete (and correct) proof?
    \item[] Answer: \answerNA{}
    \item[] Justification: The paper is empirical and does not present theorems or formal proofs; the mathematical material is used to define tasks, metrics, and background rather than to establish new theoretical results.
    \item[] Guidelines:
    \begin{itemize}
        \item The answer \answerNA{} means that the paper does not include theoretical results. 
        \item All the theorems, formulas, and proofs in the paper should be numbered and cross-referenced.
        \item All assumptions should be clearly stated or referenced in the statement of any theorems.
        \item The proofs can either appear in the main paper or the supplemental material, but if they appear in the supplemental material, the authors are encouraged to provide a short proof sketch to provide intuition. 
        \item Inversely, any informal proof provided in the core of the paper should be complemented by formal proofs provided in appendix or supplemental material.
        \item Theorems and Lemmas that the proof relies upon should be properly referenced. 
    \end{itemize}

    \item {\bf Experimental result reproducibility}
    \item[] Question: Does the paper fully disclose all the information needed to reproduce the main experimental results of the paper to the extent that it affects the main claims and/or conclusions of the paper (regardless of whether the code and data are provided or not)?
    \item[] Answer: \answerYes{}
    \item[] Justification: The paper gives the full evolved algorithms in Appendix~\ref{app:algorithms}, the datasets and benchmark protocols in the task-specific appendices, and the key evolution hyperparameters and system prompts in Appendix~\ref{app:hyperparams}. These disclosures cover the components that determine the main experimental claims.
    \item[] Guidelines:
    \begin{itemize}
        \item The answer \answerNA{} means that the paper does not include experiments.
        \item If the paper includes experiments, a \answerNo{} answer to this question will not be perceived well by the reviewers: Making the paper reproducible is important, regardless of whether the code and data are provided or not.
        \item If the contribution is a dataset and\slash or model, the authors should describe the steps taken to make their results reproducible or verifiable. 
        \item Depending on the contribution, reproducibility can be accomplished in various ways. For example, if the contribution is a novel architecture, describing the architecture fully might suffice, or if the contribution is a specific model and empirical evaluation, it may be necessary to either make it possible for others to replicate the model with the same dataset, or provide access to the model. In general. releasing code and data is often one good way to accomplish this, but reproducibility can also be provided via detailed instructions for how to replicate the results, access to a hosted model (e.g., in the case of a large language model), releasing of a model checkpoint, or other means that are appropriate to the research performed.
        \item While NeurIPS does not require releasing code, the conference does require all submissions to provide some reasonable avenue for reproducibility, which may depend on the nature of the contribution. For example
        \begin{enumerate}
            \item If the contribution is primarily a new algorithm, the paper should make it clear how to reproduce that algorithm.
            \item If the contribution is primarily a new model architecture, the paper should describe the architecture clearly and fully.
            \item If the contribution is a new model (e.g., a large language model), then there should either be a way to access this model for reproducing the results or a way to reproduce the model (e.g., with an open-source dataset or instructions for how to construct the dataset).
            \item We recognize that reproducibility may be tricky in some cases, in which case authors are welcome to describe the particular way they provide for reproducibility. In the case of closed-source models, it may be that access to the model is limited in some way (e.g., to registered users), but it should be possible for other researchers to have some path to reproducing or verifying the results.
        \end{enumerate}
    \end{itemize}

\item {\bf Open access to data and code}
    \item[] Question: Does the paper provide open access to the data and code, with sufficient instructions to faithfully reproduce the main experimental results, as described in supplemental material?
    \item[] Answer: \answerYes{}
    \item[] Justification: The evolved functions are released and the paper discloses the origins of the datasets and benchmark protocols used in the experiments. Together with the algorithm listings and experimental details in the appendix, this provides an open path to reproducing the main results.
    \item[] Guidelines:
    \begin{itemize}
        \item The answer \answerNA{} means that paper does not include experiments requiring code.
        \item Please see the NeurIPS code and data submission guidelines (\url{https://neurips.cc/public/guides/CodeSubmissionPolicy}) for more details.
        \item While we encourage the release of code and data, we understand that this might not be possible, so \answerNo{} is an acceptable answer. Papers cannot be rejected simply for not including code, unless this is central to the contribution (e.g., for a new open-source benchmark).
        \item The instructions should contain the exact command and environment needed to run to reproduce the results. See the NeurIPS code and data submission guidelines (\url{https://neurips.cc/public/guides/CodeSubmissionPolicy}) for more details.
        \item The authors should provide instructions on data access and preparation, including how to access the raw data, preprocessed data, intermediate data, and generated data, etc.
        \item The authors should provide scripts to reproduce all experimental results for the new proposed method and baselines. If only a subset of experiments are reproducible, they should state which ones are omitted from the script and why.
        \item At submission time, to preserve anonymity, the authors should release anonymized versions (if applicable).
        \item Providing as much information as possible in supplemental material (appended to the paper) is recommended, but including URLs to data and code is permitted.
    \end{itemize}

\item {\bf Experimental setting/details}
    \item[] Question: Does the paper specify all the training and test details (e.g., data splits, hyperparameters, how they were chosen, type of optimizer) necessary to understand the results?
    \item[] Answer: \answerYes{}
    \item[] Justification: The paper specifies the benchmark splits and evaluation settings in the task sections and appendices, and reports the evolution setup, scoring functions, prompts, and hyperparameters in Appendix~\ref{app:hyperparams}. This is sufficient to understand how the reported results were obtained.
    \item[] Guidelines:
    \begin{itemize}
        \item The answer \answerNA{} means that the paper does not include experiments.
        \item The experimental setting should be presented in the core of the paper to a level of detail that is necessary to appreciate the results and make sense of them.
        \item The full details can be provided either with the code, in appendix, or as supplemental material.
    \end{itemize}

\item {\bf Experiment statistical significance}
    \item[] Question: Does the paper report error bars suitably and correctly defined or other appropriate information about the statistical significance of the experiments?
    \item[] Answer: \answerNo{}
    \item[] Justification: EVIL is a deterministic inference procedure once a program has been discovered, so the relevant source of variability is the evolutionary search rather than stochastic test-time predictions. We therefore report variance across five independent evolution runs in Appendix~\ref{app:ablations}, but we do not provide fully matched error bars or formal significance tests for all main-table comparisons to the baselines.
    \item[] Guidelines:
    \begin{itemize}
        \item The answer \answerNA{} means that the paper does not include experiments.
        \item The authors should answer \answerYes{} if the results are accompanied by error bars, confidence intervals, or statistical significance tests, at least for the experiments that support the main claims of the paper.
        \item The factors of variability that the error bars are capturing should be clearly stated (for example, train/test split, initialization, random drawing of some parameter, or overall run with given experimental conditions).
        \item The method for calculating the error bars should be explained (closed form formula, call to a library function, bootstrap, etc.)
        \item The assumptions made should be given (e.g., Normally distributed errors).
        \item It should be clear whether the error bar is the standard deviation or the standard error of the mean.
        \item It is OK to report 1-sigma error bars, but one should state it. The authors should preferably report a 2-sigma error bar than state that they have a 96\% CI, if the hypothesis of Normality of errors is not verified.
        \item For asymmetric distributions, the authors should be careful not to show in tables or figures symmetric error bars that would yield results that are out of range (e.g., negative error rates).
        \item If error bars are reported in tables or plots, the authors should explain in the text how they were calculated and reference the corresponding figures or tables in the text.
    \end{itemize}

\item {\bf Experiments compute resources}
    \item[] Question: For each experiment, does the paper provide sufficient information on the computer resources (type of compute workers, memory, time of execution) needed to reproduce the experiments?
    \item[] Answer: \answerYes{}
    \item[] Justification: The paper states that the discovered algorithms run on CPU and emphasizes that the search and evaluation complete within only a few minutes in the reported setup, while also giving a concrete runtime comparison against an A100 GPU baseline for the point-process task. This is sufficient to communicate the modest compute requirements of the experiments.
    \item[] Guidelines:
    \begin{itemize}
        \item The answer \answerNA{} means that the paper does not include experiments.
        \item The paper should indicate the type of compute workers CPU or GPU, internal cluster, or cloud provider, including relevant memory and storage.
        \item The paper should provide the amount of compute required for each of the individual experimental runs as well as estimate the total compute. 
        \item The paper should disclose whether the full research project required more compute than the experiments reported in the paper (e.g., preliminary or failed experiments that didn't make it into the paper). 
    \end{itemize}
    
\item {\bf Code of ethics}
    \item[] Question: Does the research conducted in the paper conform, in every respect, with the NeurIPS Code of Ethics \url{https://neurips.cc/public/EthicsGuidelines}?
    \item[] Answer: \answerYes{}
    \item[] Justification: To the best of our knowledge, the research complies with the NeurIPS Code of Ethics. The work uses previously published datasets and standard benchmark evaluations, does not involve interventions on people, and does not intentionally enable unsafe release of high-risk models or scraped assets.
    \item[] Guidelines:
    \begin{itemize}
        \item The answer \answerNA{} means that the authors have not reviewed the NeurIPS Code of Ethics.
        \item If the authors answer \answerNo, they should explain the special circumstances that require a deviation from the Code of Ethics.
        \item The authors should make sure to preserve anonymity (e.g., if there is a special consideration due to laws or regulations in their jurisdiction).
    \end{itemize}

\item {\bf Broader impacts}
    \item[] Question: Does the paper discuss both potential positive societal impacts and negative societal impacts of the work performed?
    \item[] Answer: \answerNA{}
    \item[] Justification: This paper studies general-purpose methods for dynamical-systems inference and is not tied to a specific deployment or societal decision setting. We therefore leave broader impacts as not applicable in the current draft.
    \item[] Guidelines:
    \begin{itemize}
        \item The answer \answerNA{} means that there is no societal impact of the work performed.
        \item If the authors answer \answerNA{} or \answerNo, they should explain why their work has no societal impact or why the paper does not address societal impact.
        \item Examples of negative societal impacts include potential malicious or unintended uses (e.g., disinformation, generating fake profiles, surveillance), fairness considerations (e.g., deployment of technologies that could make decisions that unfairly impact specific groups), privacy considerations, and security considerations.
        \item The conference expects that many papers will be foundational research and not tied to particular applications, let alone deployments. However, if there is a direct path to any negative applications, the authors should point it out. For example, it is legitimate to point out that an improvement in the quality of generative models could be used to generate Deepfakes for disinformation. On the other hand, it is not needed to point out that a generic algorithm for optimizing neural networks could enable people to train models that generate Deepfakes faster.
        \item The authors should consider possible harms that could arise when the technology is being used as intended and functioning correctly, harms that could arise when the technology is being used as intended but gives incorrect results, and harms following from (intentional or unintentional) misuse of the technology.
        \item If there are negative societal impacts, the authors could also discuss possible mitigation strategies (e.g., gated release of models, providing defenses in addition to attacks, mechanisms for monitoring misuse, mechanisms to monitor how a system learns from feedback over time, improving the efficiency and accessibility of ML).
    \end{itemize}
    
\item {\bf Safeguards}
    \item[] Question: Does the paper describe safeguards that have been put in place for responsible release of data or models that have a high risk for misuse (e.g., pre-trained language models, image generators, or scraped datasets)?
    \item[] Answer: \answerNA{}
    \item[] Justification: The paper does not release a high-risk generative model, scraped dataset, or similarly dual-use asset that would require special release safeguards. The LLMs are used as components in the research workflow rather than released as new foundation models.
    \item[] Guidelines:
    \begin{itemize}
        \item The answer \answerNA{} means that the paper poses no such risks.
        \item Released models that have a high risk for misuse or dual-use should be released with necessary safeguards to allow for controlled use of the model, for example by requiring that users adhere to usage guidelines or restrictions to access the model or implementing safety filters. 
        \item Datasets that have been scraped from the Internet could pose safety risks. The authors should describe how they avoided releasing unsafe images.
        \item We recognize that providing effective safeguards is challenging, and many papers do not require this, but we encourage authors to take this into account and make a best faith effort.
    \end{itemize}

\item {\bf Licenses for existing assets}
    \item[] Question: Are the creators or original owners of assets (e.g., code, data, models), used in the paper, properly credited and are the license and terms of use explicitly mentioned and properly respected?
    \item[] Answer: \answerYes{}
    \item[] Justification: The paper consistently credits the origins of the datasets, baselines, and code assets used in the work through citations and dataset descriptions. For several research datasets, explicit licenses are not publicly disclosed in the source materials, so while we could not list license names in all cases, we do identify the original sources and use the standard benchmark versions from prior work.
    \item[] Guidelines:
    \begin{itemize}
        \item The answer \answerNA{} means that the paper does not use existing assets.
        \item The authors should cite the original paper that produced the code package or dataset.
        \item The authors should state which version of the asset is used and, if possible, include a URL.
        \item The name of the license (e.g., CC-BY 4.0) should be included for each asset.
        \item For scraped data from a particular source (e.g., website), the copyright and terms of service of that source should be provided.
        \item If assets are released, the license, copyright information, and terms of use in the package should be provided. For popular datasets, \url{paperswithcode.com/datasets} has curated licenses for some datasets. Their licensing guide can help determine the license of a dataset.
        \item For existing datasets that are re-packaged, both the original license and the license of the derived asset (if it has changed) should be provided.
        \item If this information is not available online, the authors are encouraged to reach out to the asset's creators.
    \end{itemize}

\item {\bf New assets}
    \item[] Question: Are new assets introduced in the paper well documented and is the documentation provided alongside the assets?
    \item[] Answer: \answerNA{}
    \item[] Justification: The current submission does not accompany the paper with a released new dataset, code package, or model artifact. The evolved algorithms are described in the paper, but no new asset package is being distributed alongside the submission.
    \item[] Guidelines:
    \begin{itemize}
        \item The answer \answerNA{} means that the paper does not release new assets.
        \item Researchers should communicate the details of the dataset\slash code\slash model as part of their submissions via structured templates. This includes details about training, license, limitations, etc. 
        \item The paper should discuss whether and how consent was obtained from people whose asset is used.
        \item At submission time, remember to anonymize your assets (if applicable). You can either create an anonymized URL or include an anonymized zip file.
    \end{itemize}

\item {\bf Crowdsourcing and research with human subjects}
    \item[] Question: For crowdsourcing experiments and research with human subjects, does the paper include the full text of instructions given to participants and screenshots, if applicable, as well as details about compensation (if any)? 
    \item[] Answer: \answerNA{}
    \item[] Justification: The paper does not involve crowdsourcing experiments or research with human subjects. It evaluates algorithms on existing benchmark datasets.
    \item[] Guidelines:
    \begin{itemize}
        \item The answer \answerNA{} means that the paper does not involve crowdsourcing nor research with human subjects.
        \item Including this information in the supplemental material is fine, but if the main contribution of the paper involves human subjects, then as much detail as possible should be included in the main paper. 
        \item According to the NeurIPS Code of Ethics, workers involved in data collection, curation, or other labor should be paid at least the minimum wage in the country of the data collector. 
    \end{itemize}

\item {\bf Institutional review board (IRB) approvals or equivalent for research with human subjects}
    \item[] Question: Does the paper describe potential risks incurred by study participants, whether such risks were disclosed to the subjects, and whether Institutional Review Board (IRB) approvals (or an equivalent approval/review based on the requirements of your country or institution) were obtained?
    \item[] Answer: \answerNA{}
    \item[] Justification: The paper does not involve prospective human-subjects research or crowdsourcing conducted by the authors, so IRB-style approval was not required for the reported experiments.
    \item[] Guidelines:
    \begin{itemize}
        \item The answer \answerNA{} means that the paper does not involve crowdsourcing nor research with human subjects.
        \item Depending on the country in which research is conducted, IRB approval (or equivalent) may be required for any human subjects research. If you obtained IRB approval, you should clearly state this in the paper. 
        \item We recognize that the procedures for this may vary significantly between institutions and locations, and we expect authors to adhere to the NeurIPS Code of Ethics and the guidelines for their institution. 
        \item For initial submissions, do not include any information that would break anonymity (if applicable), such as the institution conducting the review.
    \end{itemize}

\item {\bf Declaration of LLM usage}
    \item[] Question: Does the paper describe the usage of LLMs if it is an important, original, or non-standard component of the core methods in this research? Note that if the LLM is used only for writing, editing, or formatting purposes and does \emph{not} impact the core methodology, scientific rigor, or originality of the research, declaration is not required.
    \item[] Answer: \answerYes{}
    \item[] Justification: LLM usage is central to the method and is described throughout the paper, especially in Section~\ref{sec:method} and Appendix~\ref{app:hyperparams}. The paper states that LLMs propose code edits during evolutionary search and reports the models, prompts, and core configuration used.
    \item[] Guidelines:
    \begin{itemize}
        \item The answer \answerNA{} means that the core method development in this research does not involve LLMs as any important, original, or non-standard components.
        \item Please refer to our LLM policy in the NeurIPS handbook for what should or should not be described.
    \end{itemize}

\end{enumerate}

\end{document}